\documentclass[acmtocl,acmnow]{acmtrans2m}

\newtheorem{theorem}{Theorem}[section]

\newtheorem{corollary}[theorem]{Corollary}
\newtheorem{proposition}[theorem]{Proposition}

\newdef{definition}[theorem]{Definition}
\newdef{remark}[theorem]{Remark}

\title{Splitting an operator:\\
Algebraic modularity results for logics with fixpoint semantics}
            
\author{JOOST VENNEKENS, DAVID GILIS and MARC DENECKER\\
 K.U. Leuven}
            
\begin{abstract} 
 It is well known that, under certain conditions, it is possible to {\em split} logic programs under stable model semantics, i.e.~to divide such a program into a number of different ``levels'', such that the models of the entire program can be constructed by incrementally constructing models for each level.  Similar results exist for other non-monotonic formalisms, such as auto-epistemic logic and default logic.
In this work, we present a general, algebraic splitting theory for logics with a fixpoint semantics.  Together with the framework of {\em approximation theory}, a general fixpoint theory for arbitrary operators, this gives us a uniform and powerful way of deriving splitting results for each logic with a fixpoint semantics.  We demonstrate the usefulness of these results, by generalizing existing results for logic programming, auto-epistemic logic and default logic.
\end{abstract}

            
            

\usepackage{amsmath}
\usepackage{amsfonts}

\usepackage[all]{xypic}

\newcommand{\default}[3]{\frac{#1 : #2}{#3}}
\newcommand{\true}{{\bf t}}

\newcommand{\dtp}[1]{\D_{#1}^u}
\newcommand{\lowsop}[1]{C_{#1}^\downarrow}
\newcommand{\upsop}[1]{C_{#1}^\uparrow}

\newcommand{\curly}[1]{\mathcal{#1}}

\newcommand{\W}{\curly{W}}
\newcommand{\B}{\curly{B}_\Sigma}

\newcommand{\C}{\curly{C}}

\newcommand{\T}{\curly{T}}
\newcommand{\pt}{H}

\renewcommand{\t}{\tilde}

\newcommand{\D}{\curly{D}}

\newcommand{\I}{\curly{I}}

\newcommand{\Dt}[1][{}]{ \tilde {\curly{D}}_{T_{#1}} }
\newcommand{\Wt}{\tilde \W_\Sigma}
\newcommand{\Qt}{\tilde Q}
\newcommand{\res}[2]{{{#1}\rvert_{#2}}}
\newcommand{\saet}{Let $T$ be a stratifiable auto-epistemic theory. }

\renewcommand{\H}{\mathcal{H}}
\newcommand{\ext}{\sqcup}
\newcommand{\kb}{\overline{\kappa}}

\newcommand{\nf}[1]{[{#1}]}
\newcommand{\lval}[2]{{#1}\langle{#2}\rangle}

\newcommand{\lsim}[1]{\overset{{#1}}{\rightsquigarrow}}
\newcommand{\sk}{\overline{k}}
\newcommand{\Bt}{{{\tilde {\curly{B}}}_\Sigma}}
\renewcommand{\L}{\curly{L}}

\begin{document}
            
\begin{bottomstuff} 
Author's address: Joost Vennekens, Department of Computer Science, K.U. Leuven,  Celestijnenlaan 200A, B-3001 Leuven, Belgium. Email:  joost.vennekens@cs.kuleuven.ac.be.
\\Marc Denecker, Department of Computer Science, K.U. Leuven,  Celestijnenlaan 200A, B-3001 Leuven, Belgium. Email: marc.denecker@cs.kuleuven.ac.be.
\end{bottomstuff}
            
\maketitle

\section{Introduction}
\label{intro}

An important aspect of human reasoning is that it is often incremental in nature.  When dealing with a complex domain, we tend to initially restrict ourselves to  a small subset of all relevant concepts.  Once these ``basic'' concepts have been figured out, we then build another, more ``advanced'', layer of concepts on this knowledge.
%, by using our previously obtained results in reasoning about another, more ``advanced'', layer of concepts.  This process is repeated until, finally, we get a clear picture of the entire domain. 
%This way of reasoning is prevalent in all of science, where it often takes the form of moving from concrete concepts to more abstract ones, as well as in every day reasoning.
A quite illustrative example of this can be found in most textbooks on computer networking.  These typically present a seven-layered model of the way in which computers communicate.  First, in the so-called physical layer, there is only talk of hardware and concepts such as wires, cables and electronic pulses.  Once these low-level issues have been dealt with, the resulting knowledge becomes a {\em fixed} base, upon which a new layer, the data-link layer, is built.  This no longer considers wires and cables and so on, but rather talks about packages of information travelling from one computer to another.  Once again, after the workings of this layer have been figured out, this information is ``taken for granted'' and becomes part of the foundation upon which a new layer is built.  This process continues all the way up to a seventh layer, the application layer, and together all of these layers  describe the operation of the entire system.

%Of course, this process crucially depends on the assumption that concepts from higher layers do not have any influence on lower level concepts.  Indeed, if they did, it would  no longer be possible to first figure out all the low level concepts, before moving on to considering the high level ones.

In this paper, we investigate a formal equivalent of this method.  More specifically, we address the question of whether a formal theory in some non-monotonic language can be {\em split} into a number of different levels or {\em strata}, such that the formal semantics of the entire theory can be constructed by succesively constructing the semantics of the various strata. (We use the terms ``stratification'' and ``splitting'' interchangeably to denote a division  into a number of different levels.  This is a more general use of both these terms, than in literature such as  \citeN{Apt88}  and \citeN{Gelfond87}.)
Such stratifications are interesting  from both a theoretical, knowledge representational and a more practical point of view.  

On the more theoretical side, stratification results provide crucial
insight into the formal and informal semantics of a language, and hence
in its use for knowledge representation. Indeed, the human brain seems
unsuited for holding large chunks of unstructured information. When the complexity of a domain increases, we rely on
our ability to understand and describe parts of the domain and construct
a description of the whole domain by composing the descriptions of its
components. Large theories which cannot be understood as somehow being a composition of components, simply cannot be understood by
humans. Stratification results are, therefore, important, especially in the context
of nonmonotonic languages in which adding a new expression to a theory
might affects the meaning of what was already represented. Our
results will present cases where adding a new expression is guaranteed
{\em not} to alter the meaning of existing theories.  

%  In order be able to build and maintain large knowledge
% bases, it is, for instance, crucial to know which parts of a theory
% can be analysed or constructed independently.  Conversely, it is of
% course also useful to know whether combining several (independently)
% correct theories will have any unexpected side-effects.  To give just
% one illustration of this, in the case of logic programming,
% \cite{dix95} introduced a property called the {\em principle of
% modularity} as one of the fundamental properties a ``good'' semantics
% for logic programs should satisfy.  This principle is quite (but not
% completely\footnote{While Dix's principle of modularity allows the
% propogation of certain truth-values within a program, it does not
% allow one to subsequently ignore the part of the program used to
% determine these truth-values.  In other words, lower stata are never
% forgotten.}) similar to demanding that every program should admit
% certain kinds of stratifications and as such demonstrates that
% concepts such as stratificition are generally considered to be quite
% important.

On the more practical side, computing models of a theory by
incrementally constructing models of each of its levels, might offer
considerable computational gain.  Indeed, suppose that, normally, it
takes $t(n)$ time to construct the model(s) of a theory of size $n$.
If we were able to split such a theory into, say, $m$ smaller theories
of equal size $n/m$, we could use this stratification to compute the
model(s) of the theory in $m\cdot t(n/m)$ time.  As model generation
is typically quite hard, i.e.~$t(n)$ is a large function of $n$, this
could provide quite a substantial improvement.
Of course, much depends of the value of $m$.  Indeed, in the worst
case, the theory would allow only the trivial stratification in which
the entire theory is a single level, i.e.~$m = 1$, which obviously
does not lead to any gain.  However, because (as argued above)
human knowlegde  tends to exhibit a more modular structure, we
would expect real knowledge bases to be rather well-behaved in this
respect.

It is therefore not surprising that stratifiability and related
concepts, such as e.g. Dix's notion of ``modularity'' \cite{dix95}, have
already been intensively studied.
It is therefore not surprising that this issue has already been intensively studied.  Indeed, splitting results have been proven for auto-epistemic logic under the semantics of expansions \cite{aelsplit,niemela94} default logic under the semantics of extensions \cite{Tur96a} and various kinds of logic programs under the stable model semantics \cite{LT94,erdogan04,eiter97}.  In all of these works, stratification is seen as a syntactical property of a theory in a certain language under a certain formal semantics.

In this work, we take a different approach to studying this topic. 
The semantics of several (non-monotonic) logics %, in particular non-monotonic reasoning 
%formalisms,  
can be expressed  through fixpoint characterizations in 
some lattice of semantic structures.
%In such a semantics, the  meaning of a theory is described by an operator, which  revises proposed ``states of affairs''.  The  models of a theory are those states which  no longer have to be revised.  Knowing such a revision operator for a theory, should suffice to know whether it is stratifiable: this will be the case if no higher levels are ever used to revise the state of affairs for lower-level concepts.  This motivates us to
We will study the stratification of these semantical operators themselves.  As such, we are able to develop a general theory of stratification at an abstract, algebraic level and apply its results to each formalism which has a fixpoint semantics. 

This approach is especially powerful when combined with the framework of {\em approximation theory}, a general fixpoint theory for arbitrary operators, which has already proved highly useful in the study of non-monotonic reasoning.  It naturally captures, for instance, (most of) the common semantics of logic programming \cite{DMT00a},  auto-epistemic logic \cite{DMT03} and default logic \cite{DMT03}.  As such,  studying stratification within this framework, allows our abstract results to be directly and easily applicable to logic programming, auto-epistemic logic and default logic.

To make this a bit more concrete, we will now briefly sketch our method of derving splitting results.   Approximation theory defines a family of different kinds of fixpoints of operators and shows that, using a suitable class of  operators, these fixpoints correspond to a family of semantics for a number of different logics.  We will introduce the concept of a {\em stratifiable operator} and prove that such operators can be split into a number of smaller {\em component operators}, in such a way that the different kinds of fixpoints of the original operator can be constructed by incrementally constructing the corresponding fixpoints of its component operators.  These algebraic results will then be used to derive concrete splitting results for logic programming, auto-epistemic logic and default logic.  To do this, we will follow these two steps: 
\begin{longitem}
\item Firstly, we have to determine {\em syntactical} conditions which suffice to ensure that every operator corresponding to a theory which satisfies these conditions, is in fact a stratifiable operator.  This tells us that the models of such a theory under various semantics, i.e.~the various kinds of fixpoints of the associated operator, can be constructed by incrementally constructing the corresponding fixpoints of the components of this operator. 
\item Secondly, we also need to provide a precise, computable characterization of the components of stratifiable operators.  This will be done by presenting a {\em syntactical} method of deriving a number of smaller theories from the original theory and showing that the components of the original operator are precisely the operators associated with these new theories.
\end{longitem}
So, in other words, using the algebraic characterization of the semantics of a number of different logics by approximation theory, our algebraic results show how splitting can be done on a semantical level, and deriving concrete splitting results for a specific logic simply boils down to determining which syntactical notions correspond to our semantical splitting concepts.

%Its  generality is the main difference between our work and previous  works concering stratification and splitting by its generality.  Indeed, in literature such as \cite{LT94}, \cite{Tur96a} and \cite{aelsplit}, stratifiability is seen as a {\em syntactical} property of theories in a {\em certain} formalism under a {\em certain} formal semantics.  For instance, \cite{LT94} solely discusses the splitting of logic programs under the stable model semantics. 

Studying stratification at this more {\em semantical} level has three distinct advantages.  First of all, it avoids duplication of effort, as the same algebraic theory takes care of stratification in logic programming, auto-epistemic logic, default logic and indeed any logic with a fixpoint semantics.  Secondly, our results can be used to easily extend existing results to other (fixpoint) semantics of the aforementioned languages.  Finally, our work also offers greater insight into the general principles underlying various known stratification results, as we are able to study this issue in itself, free of being restricted to a particular syntax or semantics.

This paper is structured in the following way.   In Section \ref{prel}, some basic notions from lattice theory are introduced and a brief introduction to the main concepts of approximation theory is given.  Section \ref{mainsec} is the main part of this work, in which we present our algebraic theory of stratifiable operators.  In Section \ref{apps}, we then show how these abstract results can be applied to logic programming,  auto-epistemic logic and default logic, thus proving generalizations of existing results.  Parts of this paper have been published as \citeN{iclp},  which contains the results concerning logic programming, and \citeN{nmr}, which contains a shorter version of the results for auto-epistemic logic.

\section{Preliminaries}
\label{prel}

In this section, we introduce some basic concepts which will be needed later on.  Section \ref{opandlat} summarizes a number of well known definitions and results from lattice theory, while Section \ref{approxintro} contains an overview of the framework of {\em approximation theory}.

\subsection{Orders and lattices}
\label{opandlat}

A binary relation $\leq$ on a set $S$ is a {\em partial order} if it is reflexive, transitive and anti-symmetric. An element $x \in S$ is a {\em central element} if it is comparable to each other element of $S$, i.e. $x \leq y$ or $ x \geq y$ for each $y \in S$.  For each subset $R$ of $S$, an element $l$ of $S$, such that $l \leq r$ for all $r \in R$ is a {\em lower bound} of $R$.  An element $g$ in $S$ such that $g$ is a lower bound of $R$ and for each other lower bound $l$ of $R$, $l \leq g$, is called the {\em greatest lower bound}, denoted $glb(R)$, of $R$.  Similarly, an element $u$ such that for each $r \in R$, $u \geq r$ is  an {\em upper bound} of $R$ and if one such upper bound is less or equal to each other upper bound of $R$, it is the {\em least upper bound} $lub(R)$ of $R$.  

A partial order $\leq$ on a set $S$ is {\em well-founded} if each non-empty subset $R$ of $S$ has a minimal element;  it is {\em total} if  every two elements $x,y \in S$ are comparable, i.e. $x \leq y$ or $x \geq y$.

A pair $\langle L, \leq \rangle$ is a {\em lattice} if $\leq$ is a partial order on a non-empty set $L$, such that each two elements $x,y$ of $L$ have a greatest lower bound $glb(x,y)$ and a least upper bound $lub(x,y)$.  A lattice $\langle L, \leq \rangle$ is {\em complete} if each subset $L'$ of $L$ has a greatest lower bound $glb(L')$ and least upper bound $lub(L')$.  By definition, a complete lattice has a minimal (or {\em bottom}) element $\bot$ and a maximal (or {\em top}) element $\top$.  Often, we will not explicitely mention the partial order $\leq$ of a lattice  $\langle L, \leq \rangle$ and simply speak of the lattice $L$.

An {\em  operator} is a function from a lattice to itself.  An operator $O$ on a lattice $L$ is monotone if for each $x,y \in L$, such that $x \leq y$, $O(x) \leq O(y)$.  An element $x$ in $L$ is a {\em fixpoint} of $O$ if $O(x) = x$.  We denote the set of all fixpoint of $O$ by $fp(O)$.  A fixpoint $x$ of $L$, such that for each other fixpoint $y$ of $L$, $x \leq y$, is {\em the least fixpoint} $lfp(O)$ of $O$.  It can be shown \cite{tar55} that each monotone operator on a complete lattice has a unique least fixpoint.

\subsection{Approximation theory} \label{approxintro}

Approximation theory is a general fixpoint theory for arbitrary operators, which generalizes ideas found in, among others, \citeN{BS91}, \citeN{gins88} and \citeN{fitting89}.
Our presentation of this theory is based on \citeN{DMT00a}.  However, we will introduce a slightly more general definition of approximation.  For a comparison between approximation theory and related approaches, we refer to \citeN{DMT00a} and \citeN{DMT03}.
%*** CHECK REFERENCE

Let $\langle L, \leq \rangle $ be a lattice.  An element $(x,y)$ of the square $L^2$ of the domain of such a lattice, can be seen as denoting an interval $[x,y] = \{z \in L \mid x \leq z \leq y\}$.   Using this intuition, we can derive a {\em precision} order $\leq_p$ on the set $L^2$ from the order $\leq$ on $L$: for each $x,y, x', y' \in L,
%\begin{equation*}
(x,y) \leq_p (x',y')$ iff  $x \leq x' \text{ and } y' \leq y$.
%\end{equation*}
Indeed, if $(x,y) \leq_p (x',y')$, then  $[x,y] \supseteq [x', y']$.  It can easily be shown that $\langle L^2, \leq_p\rangle$ is also a lattice, which we will call the {\em bilattice} corresponding to $L$.  Moreover, if $L$ is complete, then so is $L^2$.   As an interval $[x,x]$ contains precisely one element, namely $x$ itself, elements $(x, x)$ of $L^2$ are called {\em exact}.  The set of all exact elements of $L^2$ forms a natural embedding of $L$ in $L^2$.   A pair $(x, y)$ only corresponds to a non-empty interval if $x \leq y$.  Such pairs are called {\em consistent}.  

Approximation theory is based on the study of operators on bilattices $L^2$ which are monotone w.r.t. the precision order $\leq_p$.   Such operators are called {\em approximations}.  For an approximation $A$ and elements $x,y$ of $L$, we denote by $A^1(x,y)$ and $A^2(x,y)$ the unique elements of $L$, for which $A(x,y) = (A^1(x,y), A^2(x,y))$.  An approximation {\em approximates} an operator $O$ on $L$ if for each $x \in L$, $A(x,x)$ contains $O(x)$, i.e. $A^1(x,x) \leq O(x) \leq A^2(x,x)$.  An {\em exact} approximation is one which maps exact elements to exact elements, i.e.~$A^1(x,x) = A^2(x,x)$ for all $x \in L$.  Similarly, a {\em consistent} approximation maps consistent elements to consistent elements, i.e.~if $x \leq y$ then $A^1(x, y) \leq A^2(x,y)$.  If an approximation is not consistent, it cannot approximate any operator.  Each exact approximation is also consistent and approximates a unique operator $O$ on $L$, namely that which maps each $x \in L$ to $A^1(x, x)$.   An approximation is {\em symmetric} if for each pair $(x,y) \in L^2$, if $A(x,y) = (x', y')$ then $A(y,x) = (y', x')$.  Each symmetric approximation is also exact.

For an approximation $A$ on $L^2$, the following two operators on $L$ can be defined: the function $A^1(\cdot,y)$ maps an element $x \in L $ to $A^1(x,y)$, i.e. $A^1(\cdot,y) = \lambda x.A^1(x,y)$, and the function $A^2(x,\cdot)$ maps an element $y \in L$ to $A^2(x,y)$, i.e.  $A^2(x,\cdot) = \lambda y.A^2(x,y)$.   As all such operators are monotone, they all have a unique least fixpoint.  We define an operator $\lowsop{A}$ on $L$, which maps each $y \in L$ to $lfp(A^1(\cdot,y))$ and, similarly, an operator $\upsop{A}$, which maps each $x \in L$ to $lfp(A^2(x,\cdot))$.   $\lowsop{A}$ is called the {\em lower stable operator} of $A$, while $\upsop{A}$ is the {\em upper stable operator} of $A$.  Both these operators are anti-monotone.  Combining these two operators, the operator $\C_A$ on $L^2$ maps each pair $(x,y)$ to $(\lowsop{A}(y), \upsop{A}(x))$. This operator is called the {\em partial stable operator} of $A$.  Because the lower and upper partial stable operators $\lowsop{A}$ and $\upsop{A}$ are anti-monotone, the partial stable operator $\C_A$ is monotone.   Note that if an approximation $A$ is symmetric, its lower and upper partial stable operators will always be equal, i.e. $\lowsop{A} = \upsop{A}$.

An approximation $A$ defines a number of different fixpoints: the least fixpoint of an approximation $A$ is called its {\em Kripke-Kleene fixpoint}, fixpoints of its partial stable operator $\C_A$ are {\em stable fixpoints} and the least fixpoint of $\C_A$ is called the {\em well-founded fixpoint} of $A$.
As shown in \citeN{DMT00a} and \citeN{DMT03}, these fixpoints correspond to various semantics of logic programming, auto-epistemic logic and default logic.

Finally, it should be noted that the concept of an approximation as defined in \citeN{DMT00a} corresponds to our definition of a {\em symmetric} approximation. %We, however, do not follow this convention, as non-symmetric approximations naturally occur in our work.

% become apparant in the next sections, the importance of these operators is that they allow for a uniform and algebraic characterization of a number of different semantics for logic programming, auto-epistemic logic and default logic.   In particular, such characterizations make use of the following concepts: the least fixpoint of an approximation $A$ is called its {\em Kripke-Kleene fixpoint}, fixpoints of its partial stable operator $\C_A$ are {\em stable fixpoints} and the least fixpoint of $\C_A$ is called the {\em well-founded fixpoint} of $A$.

%Finally, it should be noted that it is quite common for approximations to be {\em symmetric}, i.e.~satisfying the property that for each pair $(x,y) \in L^2$, if $A(x,y) = (x', y')$ then $A(y,x) = (y', x')$.  Indeed, in \cite{DMT00a} and other works concerning approximation theory, this requirement has even been considered as part of the definition of an approximation.  We, however, do not follow this tradition, as non-symmetric approximations naturally occur in our work.  It is easy to show that symmetric approximations are always exact.  Moreover,if an approximation $A$ is symmetric, its lower and upper partial stable operators will always be equal, i.e. $\lowsop{A} \equiv \upsop{A}$.

\section{Stratification of operators}
\label{mainsec}

In this section, we develop a theory of stratifiable operators.  We will, in section \ref{opprodlat}, investigate operators on a special kind of lattice, namely {\em product lattices}, which will be introduced in section \ref{prodlat}.  In section \ref{approxstrat}, we  then return to approximation theory and discuss stratifiable approximations on product lattices. % Finally, in section \ref{otherops}, we will relate the results obtained for operators on product lattices to operators on other kinds of lattices.

\subsection{Product lattices}
\label{prodlat}

We begin by defining the notion of a {\em product set}, which is a generalization of the well-known concept of cartesian products. 

\begin{definition} Let $I$ be a set, which we will call the {\em index set} of the product set, and for each $i \in I$, let $S_i$ be a set.  The {\em product set} $\bigotimes_{i \in I} S_i$ is the following set of functions:
\begin{equation*}
\bigotimes_{i \in I} S_i = \{ f \mid f:I \rightarrow \bigcup_{i \in I} S_i\text{ such that } \forall i \in I: f(i) \in S_i\}.
\end{equation*}
\end{definition}

Intuitively, a product set $\bigotimes_{i \in I} S_i$ contains all ways of selecting one element from each set $S_i$.  As such, if the index set $I$ is a set with $n$ elements, e.g.~the set $\{1,\ldots,n\}$, the product set $\bigotimes_{i \in I} S_i$ is simply (isomorphic to) the cartesian product $S_1 \times \cdots \times S_n$.

\begin{definition} Let $I$ be a set and for each $i \in I$, let $\langle S_i, \leq_i \rangle$ be a partially ordered set.  The {\em product order} $\leq_\otimes$ on the set $\otimes_{i \in I} S_i$ is defined by $\forall x,y \in \bigotimes_{i \in I} S_i  :$
\begin{equation*}
x \leq_\otimes y \qquad iff \qquad \forall i \in I: x(i) \leq_i y(i).
\end{equation*}
\end{definition}

It can easily be shown that if all of the partially ordered sets $S_i$ are (complete) lattices, the product set $\otimes_{i \in I} S_i$, together with its product order $\leq_\otimes$, is also a (complete) lattice.  We therefore refer to the pair $\langle\otimes_{i \in I} S_i, \leq_\otimes \rangle$ as the {\em product lattice} of lattices $S_i$.

From now on, we will only consider product lattices with a {\em well-founded} index set, i.e.~index sets $I$ with a partial order $\preceq$  such that each non-empty subset of $I$ has a $\preceq$-minimal element.  This will allow us to use inductive arguments in dealing with elements of product lattices.  Most of our results, however, also hold for index sets with an arbitrary partial order; if a certain proof depends on the well-foundedness of $I$, we will always explicitely mention this.

In the next sections, the following notations will be used. For a function $f: A \rightarrow B$ and a subset $A'$ of $A$, we denote by $\res{f}{A'}$ the restriction of $f$ to $A'$, i.e.~$\res{f}{A'} : A' \rightarrow B: a' \mapsto f(a')$.  For an element $x$ of a product lattice $\otimes_{i \in I} L_i$ and an $i \in I$,  we abbreviate $\res{x}{\{j \in I \mid j \preceq i\}}$ by  $\res{x}{\preceq i}$.  We also use similar abbreviations $\res{x}{\prec i}$, $\res{x}{i}$ and $\res{x}{\not \preceq i}$.  If $i$ is a minimal element of the well-founded set $I$, $\res{x}{\prec i}$ is defined as the empty function.  For each index $i$, the set $\{\res{x}{\preceq i} \mid x \in L\}$, ordered by the appropriate restriction $\res{\leq_\otimes}{\preceq i}$ of the product order, is also a lattice.  Clearly, this sublattice of $L$ is isomorphic to the product lattice $\otimes_{j \preceq i} L_i$.  We denote this sublattice by $\res{L}{\preceq i}$ and use a similar notation $\res{L}{\prec i}$ for $\otimes_{j \prec i} L_i $.

If $f,g$ are functions $f: A \rightarrow B$, $g : C \rightarrow D$ and the domains $A$ and $C$ are disjoint, we denote by $f \ext g$ the function from $A \cup C$ to $B \cup D$, such that for all $a \in A$,  $(f \ext g)(a) = f(a)$ and for all $c \in C$, $(f \ext g)(c) = g(c)$.  Furthermore, for any $g$ whose domain is disjoint from the domain of $f$, we call  $f \ext g$ an {\em extension} of $f$.  For each element $x$ of a product lattice $L$ and each index $i \in I$, the extension $\res{x}{\prec i} \ext \res{x}{i}$ of  $\res{x}{\prec i}$ is clearly equal to $\res{x}{\preceq i}$.  For ease of notation, we sometimes simply write $x(i)$ instead of $\res{x}{i}$ in such expressions, i.e.~we identify an element $a$ of the $i$th lattice $L_i$ with the function from $\{i\}$ to $L_i$ which maps $i$ to $a$.  Similarly, $\res{x}{\prec i} \ext x(i) \ext \res{x}{\not \preceq i} = x$.

%In the next sections, 
We will use the symbols $x,y$ to denote elements of an entire product lattice $L$; $a,b$ to denote elements of a single level $L_i$ and $u, v$ to denote elements of $\res{L}{\prec i}$.

\subsection{Operators on product lattices}
\label{opprodlat}

Let  $\langle I, \preceq \rangle$ be a well-founded index set and let $L = \otimes_{i \in I} L_i$ be the product lattice of lattices $\langle L_i, \leq_i \rangle_{i \in I}$.
Intuitively, an operator $O$ on  $L$ is stratifiable over the order $\preceq$, if the value $(O(x))(i)$ of $O(x)$ in the $i$th stratum only depends on values $x(j)$ for which $j \preceq i$.  This is formalized in the following definition.

\begin{definition} An operator $O$ on a product lattice $L$ is {\em stratifiable} iff $\forall x,y \in L, \forall i \in I:$ if $\res{x}{\preceq i} = \res{y}{\preceq i}$ then $\res{O(x)}{\preceq i} = \res{O(y)}{\preceq i}$.
\end{definition}

It is also possible to characterize stratifiablity in a more constructive manner.  The following theorem shows that stratifiablity of an operator $O$ on a product lattice $L$ is equivalent to the existence of a family of operators on each lattice $L_i$ (one for each partial element $u$ of $\res{L}{\prec i}$), which mimics the behaviour of $O$ on this lattice.

\begin{proposition} \label{comps}  Let $O$ be an operator on a product lattice $L$.  $O$ is stratifiable iff for each $i \in I$ and $u \in \res{L}{\prec i}$ there exists a unique operator $O_i^u$ on $L_i$, such that for all $x \in L$:
\begin{equation*}\text{If } \res{x}{\prec i} = u \text{ then } (O(x))(i) =  O_i^u(x(i)).
\end{equation*}
\end{proposition}

\begin{proof} To prove the implication from left to right, let $O$ be a stratifiable operator, $i \in I$ and $u \in \res{L}{\prec i}$.  We define the operator $O^u_i$ on $L_i$ as
\begin{equation*}
 O^u_i: L_i \rightarrow L_i: a \mapsto (O(y))(i),
\end{equation*}
with $y$ some element of $L$ extending $ u \ext a$.   Because of the stratifiability of $O$, this operator is well-defined and it trivially satisfies the required condition.

To prove the other direction, suppose the right-hand side of the equivalence holds and let $x,x'$ be elements of $L$, such that $\res{x}{\preceq i} = \res{x'}{\preceq i}$.  Then for each $j \preceq i$, $(O(x))(j) =  O_j^\res{x}{\prec j}(x(j)) =  O_j^\res{x'}{\prec j} (x'(j)) = (O(x'))(j)$.
\end{proof}

The operators $O_i^u$ are called the {\em components} of $O$.   Their existence allows us to already prove one of the main theorems of this paper, which states that is possible to construct the fixpoints of a stratifiable operator in a bottom-up manner w.r.t. the well-founded order $\preceq$ on the index set.

\begin{theorem} \label{main} Let $O$ be a stratifiable operator on a product lattice $L$.  Then for each $ x \in L$:
\begin{equation*}
x \text{ is a fixpoint of } O \qquad iff \qquad \forall i \in I: x(i) \text{ is a fixpoint of } O_i^{\res{x}{\prec i}}.
\end{equation*} 
\end{theorem}
\begin{proof} Follows immedately from proposition \ref{comps}.
\end{proof}

If $O$ is a monotone operator on a complete lattice, we are often interested in its {\em least} fixpoint.  This can also be constructed by means of the least fixpoints of the components of $O$.  Such a construction of course requires each component to actually have a least fixpoint as well.  We will therefore first show that the components of a monotone operator are also monotone.

\begin{proposition} \label{monotonecomps} Let $O$ be a stratifiable operator on a product lattice $L$, which is monotone w.r.t.~the product order $\leq_\otimes$.  Then for each $i \in I$ and $u \in \res{L}{\prec i}$, the component $O_i^u:L_i \rightarrow L_i$ is monotone w.r.t.~to the order $\leq_i$ of the $i$th lattice $L_i$ of $L$.
\end{proposition}

\begin{proof}
Let $i$ be an index in $I$, $u$ an element of $\res{L}{\prec i}$ and $a,b$ elements of $L_i$, such that $a \leq_i b$.  Let $x,y \in L$, such that $x$ extends $u \ext a$, $y$ extends $u \ext b$ and for each $j \not \preceq i$, $x(j) = y(j)$.   Because of the definition of $\leq_\otimes$, clearly $x \leq_\otimes y$ and therefore $\forall j \in I: O_j^{\res{x}{\prec j}}(x(j)) =(O(x))(j) \leq_j (O(y))(j) = O_j^{\res{y}{\prec j}}(y(j))$, which, taking $j = i$, implies $O_i^u(a) \leq_i O_i^u(b)$.
\end{proof}

Now, we can prove that the least fixpoints of the components of a monotone stratifiable operator indeed form the least fixpoint of the operator itself.  We will do this, by first proving the following, slightly more general theorem, which we will be able to reuse later on.

\begin{proposition}  \label{monstrat}  Let $O$ be a monotone operator on a complete product lattice $L$ and let for each $i \in I$, $u \in \res{L}{\prec i}$, $P_i^u$ be a monotone operator on $L_i$ (not necessarily a component of $O$), such that:
\begin{equation*} \label{fps}
x \text{ is a fixpoint of } O \qquad iff \qquad \forall i \in I: x(i) \text{ is a fixpoint of } P_i^{\res{x}{\prec i}}.
\end{equation*} \label{lfps}
Then the following equivalence also holds:
\begin{equation*}
x \text{ is the least fixpoint of } O \qquad iff \qquad \forall i \in I: x(i) \text{ is the least fixpoint of } P_i^{\res{x}{\prec i}}.
\end{equation*}
\end{proposition}

\begin{proof}
To prove the implication from left to right, let $x$ be the least fixpoint of $O$ and let $i$ be an arbitrary index in $I$.  We will show that for each fixpoint $a$ of $P_i^\res{x}{\prec i}$, $a \geq x(i)$.  So, let $a$ be such a fixpoint.   We can inductively extend $\res{x}{\prec i} \ext a$ to an element $y$ of $L$ by defining for all $j \not \preceq i$,  $y(j)$ as $lfp(P_j^\res{y}{\prec j})$.   Because of the well-foundedness of $\preceq$, $y$ is well defined.  Furthermore, $y$ is clearly also a fixpoint of $O$. Therefore $x \leq y$ and, by definition of the product order on $L$, $x(i) \leq_i y(i) = a$.  

To prove the other direction, let $x$ be an element of $L$, such that, for each $i \in I$, $x(i)$ is the least fixpoint of $P_i^\res{x}{\prec i}$.   Now, let $y$ be the least fixpoint of $O$.    To prove that $x = y$, it suffices to show that for each $i \in I$, $\res{x}{\preceq i} = \res{y}{\preceq i}$.   We will prove this by by induction on the well-founded order $\preceq$ of $I$.  If $i$ is a minimal element of $I$, the proposition trivially holds.  Now, let $i$ be an index which is not the minimal element of $I$ and assume that for each $j \prec i$, $\res{x}{\preceq j} = \res{y}{\preceq j}$.  It suffices to show that $x(i) = y(i).$ Because $y$ is a fixpoint of $O$, $y(i)$ is fixpoint of $P_i^\res{y}{\prec i}$.   As the induction hypothesis  implies that $\res{x}{\prec i} = \res{y}{\prec i}$, $y(i)$ is a also fixpoint of   $P_i^\res{x}{\prec i}$ and therefore $x(i) \leq y(i)$.  However, because $x$ is also a fixpoint of $O$ and therefore must be greater than the least fixpoint $y$ of $O$, the definition of the product order on $L$ implies that $x(i) \geq y(i)$ as well.  Therefore $x(i) = y(i)$.
\end{proof}

It is worth noting that the condition that the order $\preceq$ on $I$ should be well-founded is necessary for this proposition to hold.  Indeed, consider for example the product lattice $L = \bigotimes_{z \in \mathbb{Z}} \{0, 1\}$, with $\mathbb{Z}$ the integers ordered by their usual, non-well-founded order.  Let $O$ be the operator mapping each $x \in L$ to the element $y : \mathbb{Z} \rightarrow \{0, 1\}$ of $L$, which maps each $z \in \mathbb{Z}$ to $0$ if $x(z - 1) = 0$ and to $1$ otherwize.   This operator is stratifiable over the order $\leq$ of $\mathbb{Z}$ and its components are the family of operators $O_z^u$, with $z \in \mathbb{Z}$ and $u \in \res{L}{\leq z}$, which are defined as mapping both 0 and 1 to 0 if $u(z - 1) = 0$ and to 1 otherwize. 
Clearly, the bottom element $\bot_L$ of $L$, which maps each $z \in \mathbb{Z}$ to 0, is the least fixpoint of $O$.  
%However, consider the family of operators $P_z^u$, with $z \in \mathbb{Z}$ and $u \in \res{L}{\leq z}$, which are defined as mapping both 0 and 1 to 0 if $u(z - 1) = 0$ and to 1 otherwize.  
However, the element $x \in L$ which maps each $z \in \mathbb{Z}$ to 1 satisfies the condition that for each  $z \in \mathbb{Z}$, $x(z)$ is the least fixpoint of $P_z^\res{x}{< z}$, but is clearly not the least fixpoint of $O$.

Together with theorem \ref{main} and proposition \ref{monotonecomps}, this proposition of course implies that for each stratifiable operator $O$ on a product lattice $L$, an element $x \in L$ is the least fixpoint of $O$ iff $\forall i \in I$, $x(i)$ is the least fixpoint of $ O_i^{\res{x}{\prec i}}$.  In other words, the least fixpoint of a stratifiable operator can also be incrementally constructed.

%\begin{corollary} Let $O$ be a stratifiable operator on a product lattice $L$.  Then for each $ x \in L$:
%\begin{equation*}
%x \text{ is the least fixpoint of } O \qquad iff \qquad \forall i \in I: x(i) \text{ is least fixpoint of } O_i^{\res{x}{\prec i}}.
%\end{equation*}
%\end{corollary}

\subsection{Approximations on product lattices}
\label{approxstrat}

In section \ref{approxintro}, we introduced several concepts from approximation theory, pointing out that we are mainly interested in studying Kripke-Kleene,  stable and well-founded fixpoints of approximations.   Similar to our treatment of general operators in the previous section, we will in this section investigate the relation between these various fixpoints of an approximation and its components. In doing so, it will be convenient to switch to an alternative representation of the bilattice $L^2$ of a product lattice $L = \otimes_{i \in I} L_i$.  Indeed, this bilattice is clearly isomorphic to the  structure $\otimes_{i \in I} L_i^2$, i.e.~to a product lattice of bilattices.  From now on, we will not distinguish between these two representations.  More specifically, when viewing $A$ as a stratifiable operator, it will be convenient to consider its domain equal to $\otimes_{i \in I} L_i^2$, while when viewing $A$ as an approximation, the representation $(\otimes_{i \in I} L_i)^2$ is more natural.

Obviously, this isomorphism and the results of the previous section already provide a way of constructing the Kripke-Kleene fixpoint of a stratifiable approximation $A$, by means of its components $A_i^u$.  Also, it is clear that if $A$ is both exact and stratifiable, the unique operator $O$ approximated by $A$ is stratifiable as well.  Indeed, this is a trivial consequence of the fact that $A(x,x) = (O(x), O(x))$ for each $x \in L$.

These results leave only the  stable and well-founded fixpoints of $A$ to be investigated.  %This of course requires stratifying the partial stable operator $\C_A$ of $A$.  %The most natural way of doing so, is by following the same process as by which this operator is usually defined.   
We will %therefore 
first examine the operators  $A^1(\cdot, y)$ and $A^2(x, \cdot)$, and then move on to the lower and upper stable operators $\lowsop{A}$ and $\upsop{A}$, before finally getting to the partial stable operator $\C_A$ itself.

\begin{proposition} \label{A1} Let $L$ be a product lattice and let $A:  L^2 \rightarrow  L^2$ be a stratifiable  approximation.  Then, for each $x, y \in L$, the operators $A^1(\cdot, y)$,  $A^2(x, \cdot)$ are also stratifiable.
Moreover, for each $i \in I$, $u \in \res{L}{\prec i}$, the components of these operators are:
\begin{align*}
(A^1(\cdot,y))^u_i = &(A_i^{(u, \res{y}{\prec i})})^1(\cdot, y(i));\\
(A^2(x,\cdot))^u_i = &(A_i^{(\res{x}{\prec i}, u)})^2(x(i),\cdot).
\end{align*}
\end{proposition}

\begin{proof}  Let $x,y$ be elements of $L$, $i$ an element of $I$. Then, because $A$ is stratifiable, $(A(x,y))(i) = (A^\res{(x,y)}{\prec i}_i) (x(i), y(i))$.  From this, the two equalities follow.
\end{proof}

In the previous section, we showed that the components of a monotone operator are monotone as well (proposition \ref{monotonecomps}).  This result obviously implies that the components $A_i^u$ of a stratifiable approximation are also approximations. Therefore, such a component $A_i^u$ itself  has a lower and uppers stable operator $\lowsop{A_i^u}$ and $\upsop{A_i^u}$ as well.   It turns out that the lower and upper stable operators of the components of $A$, characterize the components of the lower and upper stable operators of $A$.

\begin{proposition} \label{lowup}  Let $L$ be a product lattice and let $A$ be a stratifiable  approximation on $L^2$. Then the operators $\lowsop{A}$ and  $\upsop{A}$ are also stratifiable.  Moreover, for each $x,y \in L$, 
\begin{align*}
 x = \lowsop{A}(y) \qquad \text{ iff }&\qquad \text{for each }i \in I, x(i) = \lowsop{A_i^\res{(x,y)}{\prec i}}(y(i)); \\
y = \upsop{A}(x)\qquad \text{ iff }& \qquad \text{for each }i \in I, y(i) = \upsop{A_i^\res{(x,y)}{\prec i}}(x(i)).
\end{align*}
\end{proposition}

\begin{proof}
Let $x, y$ be elements of $L$.  Because $A^1(\cdot, y)$ is stratifiable (proposition \ref{A1}), the corollary to proposition \ref{monstrat} implies that $x = \lowsop{A}(y) = lfp(A^1(\cdot,y))$ iff for each $i \in I$, $x(i) = lfp\big((A^1(\cdot,y))_i^{\res{x}{\prec i}}\big)$.  Because of proposition \ref{A1}, this is in turn equivalent to for each $i \in I$,  $x(i) = lfp\big((A_i^{\res{(x,y)}{\prec i}})^1(\cdot,y(i))\big) = \lowsop{A_i^{\res{(x,y)}{\prec i}}}(y(i))$.  The proof of the second equivalence is similar.
\end{proof}

This  proposition shows how, for each $x,y \in L$, $\lowsop{A}(y)$ and $\upsop{A}(x)$ can be be constructed  incrementally from the upper and lower stable operators corresponding to the components of $A$.  This result also implies a similar property for the partial stable operator $\C_A$ of an approximation $A$.

\begin{proposition} \label{CA} Let $L$ be a product lattice and let $A:  L^2 \rightarrow  L^2$ be a stratifiable  approximation.  Then the operator $\C_A$ is also stratifiable.  Moreover, for each $x,x',y,y' \in L$, the following equivalence holds:
\begin{equation*}
(x', y') = \C_A(x,y) \qquad iff \qquad \forall i \in I: \left\{\begin{aligned} x'(i) =& \lowsop{A_i^\res{(x',y)}{\prec i}}(y(i));\\ y'(i) =& \upsop{A_i^\res{(x,y')}{\prec i}}(x(i)).\end{aligned}\right.
\end{equation*}
\end{proposition}

\begin{proof} The above equivalence follows immediately from proposition \ref{lowup}.  To prove the stratifiability of $\C_A$, suppose that $x_1, y_1, x_2, y_2 \in L$, such that $\res{(x_1, y_1)}{\preceq i} = \res{(x_2, y_2)}{\preceq i}$.   Let $(x_1', y_1') = \C_A(x_1, y_1)$ and $(x_2', y_2') = \C_A(x_2, y_2)$.  It suffices to show that $\forall j \preceq i$, $x_1'(j) = x_2'(j)$ and $y_1'(j) = y_2'(j)$.  We show this by induction on $\preceq$.   Firstly, if $j$ is minimal, then 
$\lowsop{A_j}(y_1(j)) = \lowsop{A_j}(y_2(j))$ and  $\upsop{A_j}(x_1(j)) = \upsop{A_j}(x_2(j))$.
Secondly, if $j$ is not minimal, then $\lowsop{A_j^{\res{(x_1',y_1)}{\prec j}}}(y_1(j)) = \lowsop{A_j^{\res{(x_2',y_2)}{\prec j}}}(y_2(j))$ and $\upsop{A_j^{\res{(x_1,y_1')}{\prec j}}}(x_1(j)) = \upsop{A_j^{\res{(x_2,y_2')}{\prec j}}}(x_2(j))$, because obviously $\res{y_1}{\preceq j} = \res{y_2}{\preceq j}$ and $\res{x_1}{\preceq j} = \res{x_2}{\preceq j}$, while  the induction hypothesis states that $\res{x_1'}{\prec j} = \res{x_2'}{\prec j}$ and $\res{y_1'}{\prec j} = \res{y_2'}{\prec j}$.
\end{proof}

It should be noted that the components $(\C_A)_i^{(u,v)}$ of the partial stable operator of a stratifiable approximation $A$ are (in general) not equal to the partial stable operators $\C_{A_i^{(u,v)}}$  of the components of $A$. Indeed, $(\C_A)_i^{(u,v)} = ((\lowsop{A})_i^v, (\upsop{A})_i^u))$, whereas $\C_{A_i^{(u,v)}} = (\lowsop{A_i^{(u,v)}}, \upsop{A_i^{(u,v)}})$.  Clearly, these two pairs are, in general, not equal, as $(\lowsop{A})_i^v$ ignores the argument $u$, which does appear in $\lowsop{A_i^{(u,v)}}$.  We can, however, characterize the fixpoints of $\C_A$, i.e.~the partial stable fixpoints of $A$,  by means of the partial stable fixpoints of the components of $A$. 

\begin{theorem} \label{approxmain} Let $L$ be a product lattice and let $A:  L^2 \rightarrow  L^2$ be a stratifiable  approximation.  Then for each element $(x,y)$ of $L^2$:
\begin{equation*}
(x,y) \text{ is a fixpoint of }\C_A \qquad iff \qquad \forall i \in I: (x,y)(i)\text{ is a fixpoint of }\C_{A_i^\res{(x,y)}{\prec i}}.
\end{equation*}

\end{theorem}
\begin{proof}  Let $x, y$ be elements of $L$, such that $(x,y) = \C_A(x,y)$.  By proposition \ref{CA}, this is equivalent to for each $i \in I$,  $x = \lowsop{A_i^\res{(x,y)}{\prec i}}(y(i))$ and $y = \upsop{A_i^\res{(x,y)}{\prec i}}(x(i))$.
\end{proof}

By proposition \ref{monstrat}, this theorem  has the following corollary:

\begin{corollary} Let $L$ be a product lattice and let $A:  L^2 \rightarrow  L^2$ be a stratifiable  approximation.
  Then for each element $(x,y)$ of $L^2$:
\begin{equation*}
(x,y) = lfp(\C_A) \qquad \text{ iff } \qquad \forall i \in I: (x,y)(i) = lfp(\C_{A_i^\res{(x,y)}{\prec i}}).
\end{equation*}

\end{corollary}

Putting all of this together, the main results of this section can be summarized as follows. If $A$ is a stratifiable approximation on a product lattice $L$, then a pair $(x,y)$ is a fixpoint, Kripke-Kleene fixpoint,  stable fixpoint or well-founded fixpoint of $A$ iff for each $i \in I$, $(x(i), y(i))$ is a fixpoint, Kripke-Kleene fixpoint, stable fixpoint or well-founded fixpoint of the component $A_i^\res{(x,y)}{\prec i}$ of $A$.  Moreover, if $A$ is  exact then an element $x \in L$ is a fixpoint of the unique operator $O$  approximated by $A$ iff for each $i \in I$, $(x(i), x(i))$  is a fixpoint of the component $A_i^\res{(x,x)}{\prec i}$ of $A$.  These characterizations give us a way of  incrementally constructing each of these fixpoints.

\subsection{Operators on other lattices}
\label{otherops}

The theory developed in the previous sections allows us to incrementally construct fixpoints of  operators on  product lattices.  However, not every operator is (isomorphic to) an operator on a (non-trivial) product lattice.  For instance, as we will see in Section \ref{ael}, the $\D_T$-operator of auto-epistemic logic is not. 
 So, the question rises whether it is also possible to incrementally construct the fixpoints of an operator $O$ on a lattice $L$, which is {\em not} a product lattice.  Clearly, this could be done if the fixpoints of $O$ were uniquely determined by the fixpoints of some other operator $\t O$ on a lattice $\t L$, where $\t L$ {\em is} a product lattice.

Therefore, this section will investigate similarity conditions which suffice to ensure the existence of such a correspondence between the fixpoints $fp(\t O)$ of an operator $\t O$ on $\t L$ and those of an operator $O$ on $L$.  More precisely, we will determine a number of properties for a function $k$ from $\t L$ to $L$, such that applying $k$ to the fixpoints of $\t O$ yields the fixpoints of $O$. For a set $S$, function $f: S \rightarrow T$ and $t \in T$, we denote the set $\{f(s) \mid s \in S\}$ by $f(S)$ and the set $\{s \in S \mid f(s) = t\}$ by $f^{-1}(t)$.  Using these notations, we can rephrase our goal as determining conditions which ensure that $k(fp(\t O)) = fp(O)$. 

A natural condition to impose on such a function $k$ from $\t L$ to $L$ is that $k \circ \t O$ should be equal to $O \circ k$.  

\begin{proposition} \label{ool1} Let  $\t L$ and  $L$ be lattices, $\t O$ an operator on $\t L$ and $O$ an operator on $L$.  If there exists a function $k$ from $\t L$ to $L$, such that $k \circ \t O = O \circ k$, then 
\begin{equation*}
k(fp(\t O)) \subseteq fp(O).
\end{equation*}
Moreover, for each fixpoint $x$ of $O$, $k^{-1}(x)$ is closed under application of $\t O$, i.e.~for each $\t x \in k^{-1}(x)$, $\t O(\t x)) \in k^{-1}(x)$.
\end{proposition}
\begin{proof}  Let $\t L, L, \t O, O$ and $k$ be as above.  Let $\t x \in \t L$ be a fixpoint of $\t O$.  Then $k(\t x) = k(\t O(\t x)) = O(k(\t x))$ and therefore $k(\t x)$ is a fixpoint of $O$.  
Let $x$ be a fixpoint of $O$ and let $\t x \in  k^{-1}(x)$.  Then $k(\t O(\t x)) = O(k (\t x)) = O(x) = x$.
\end{proof}

This condition is, however, not strong enough to ensure that the reverse inclusion $k(fp(\t O)) \supseteq fp(O)$ also holds.   More precisely, there are two ways in which a fixpoint $x$ of $O$ can be in $L \setminus k(fp(\t O))$.  

First of all, $k^{-1}(x)$ may be empty.  Consider, for instance, the case in which $\t L$ is a one-element lattice $\{a\}$ with $\t O$ mapping $a$ to $a$ and $L$ is a two-element lattice $\{b,c\}$ with $O$ mapping $b$ to $b$ and $c$ to $c$.  In this case, the function $k$ mapping $a$ to $b$ satisfies the above condition, but $k(fp(\t O))$ clearly contains only $b$.   Of course, if $O(L) \subseteq k(\t L)$, i.e.~each $k^{-1}(O(x))$ is non empty, this cannot happen.  

\begin{proposition}\label{ool2} Let  $\t L$ and  $L$ be lattices, $\t O$ an operator on $\t L$ and $O$ an operator on $L$.  If there exists a function $k$ from $\t L$ to $L$, such that $k \circ \t O = O \circ k$ and $O(L) \subseteq k(\t L)$, then for each fixpoint $x$ of $O$, $k^{-1}(x) \neq \{\}$, i.e.~there exists a $\t x \in \t L$, such that $k(\t x) = x$.
\end{proposition}

Secondly, even if $k^{-1}(x)$ is not empty, this set does not necessarily contain a fixpoint of $\t O$.  Consider for instance the case in which $\t L$ is a two-element lattice $\{a, b\}$, with $\t O$ mapping $a$ to $b$ and $b$ to $a$, and $L$ a one-element lattice $\{c\}$, with $O$ mapping $c$ to itself.  Then the function $k$ mapping both $a$ and $b$ to $c$ satisfies the above condition, but $\t O$ has no fixpoints. 

To rule out such cases, we can make use of the fact that each $k^{-1}(x)$ is closed under application of $\t O$ (proposition \ref{ool1}).   Let us first assume that $\t L$ is finite.  In this case, situations in which, for some fixpoint $x$ of $O$, $k^{-1}(x)$ does not contain a fixpoint of $\t O$ can only occur if $\t O$ oscillates between some elements of  $k^{-1}(x)$.   Therefore, it suffices for the operator $\t O$ to be monotone and for each $k^{-1}(x)$ to contain at least one pair $\t x$ and $\t O(\t x)$ which are comparable.  Indeed, under such  conditions, each $k^{-1}(x)$ will contain an element $\t c$, such that successive applications of $\t O$ on this $\t c$ form a chain, i.e.~a totally ordered subset, of $k^{-1}(x)$, which is guaranteed to reach a fixpoint.

\begin{proposition}Let  $\t L$ and  $L$ be lattices, $\t O$ an operator on $\t L$ and $O$ an operator on $L$.  If there exists a function $k$ from $\t L$ to $L$, such that $k \circ \t O = O \circ k$, $O(L) \subseteq k(\t L)$ and
\begin{itemize}
\item $\t O$ is monotone,
\item  each non-empty set $k^{-1}(x)$, with $x \in L$,  has a central element (i.e.~one which is comparable to all other elements of $k^{-1}(x)$),
\item $\t L$ is finite,
\end{itemize}
then $k(fp(\t O)) = fp(O)$.
\end{proposition}
\begin{proof}
Let $\t L, \t O, L$ and $O$ be as above.  Let $x$ be a fixpoint of $O$. It suffices to show the existence of a fixpoint of $\t O$ in $k^{-1}(x)$.  Because the set $k^{-1}(x)$ is not empty (proposition \ref{ool2}), it contains a central element $\t c$.  Because $k^{-1}(x)$ is closed under application of $\t O$ (proposition \ref{ool1}), $\t O(\t c)$ is comparable to $\t c$.  As $\t O$ is monotone, this implies that the elements $(\t O^n(\t c))_{n \in \mathbb{N}}$ form a chain.  Moreover, by induction on $\mathbb{N}$, all these elements are clearly in $k^{-1}(x)$.   Because $\t L$ is finite, there must be a $m \in \mathbb{N}$, for which $\t O^m(\t c)$ is a fixpoint of $\t O$.
\end{proof}

However, this still does not quite suffice for infinite lattice.  Consider, for instance, the case in which $\t L$ is the complete lattice $\mathbb{N} \cup \{\infty\}$, with the standard order on the natural numbers and $\infty = glb(\mathbb{N})$.   Let $\t O$ be the operator on $\t L$ which maps $\infty$ to $\infty$ and each $n \in \mathbb{N}$ to $n + 1$.  Let $L$ be the two-element lattice $\{0,1\}$ and let $O$ map $0$ to $0$ and $1$ to $1$.  Then the function $k$ which maps each $n \in \mathbb{N}$ to $0$ and $\infty$ to $1$ satisfies all the above conditions and yet there is no fixpoint of $\t O$ in $k^{-1}(0)$.

As can be seen from this example, the problem is caused by the fact that (for infinite lattices) $k^{-1}(x)$ being closed under application of $\t O$, does not suffice to ensure that --- for an ordinal $\alpha$ greater or equal to the ordinality $\omega$ of the natural numbers --- the result $O^\alpha(\t x)$ of applying $\t O$ $\alpha$ times to an element $\t x \in k^{-1}(x)$ will still be in $k^{-1}(x)$.  As, however, even for an infinite lattice, it is still the case that elements $(\t O^\beta(\t x))_{\beta < \alpha}$ form a chain, this problem can never occur if $k$ is chain continuous, i.e.~if for each chain $(\t x_i)_{i \leq \alpha}$, $k(glb(\{\t x_i \mid i \leq \alpha\})) = glb(\{k(\t x_i) \mid i \leq \alpha\})$ and  $k(lub(\{\t x_i \mid i \leq \alpha\})) = lub(\{k(\t x_i) \mid i \leq \alpha\})$.
This motivates the following definitions.

\begin{definition} Let  $\t L$ and  $L$ be lattices.   If there exists a function $k$ from $\t L$ to $L$, such that each non-empty set $k^{-1}(x)$, with $x \in L$, has a central element and  $k$ is chain-continuous, then $\t L$ is {\em $k$-similar} to $L$.  This is denoted by $\t L \lsim{k} L$.
\end{definition}

\begin{definition} Let  $\t L$ and  $L$ be lattices, $\t O$ an operator on $\t L$ and $O$ an operator on $L$.  If there exists a function $k$ from $\t L$ to $L$, such that $k \circ \t O = O \circ k$ and $O(L) \subseteq k(\t L)$, $\t O$ {\em $k$-mimics} $O$.  This is denoted by $\t O \sim{k} O$.
\end{definition}

Now, we finally have a set of properties, which suffices to ensure correspondence between the fixpoints of $\t O$ and those of $O$.

\begin{proposition} \label{mimicsfp} Let  $\t L$ and  $L$ be complete lattices, $\t O$ an operator on $\t L$ and $O$ an operator on $L$, such that $\t L \lsim{k} L$,  $\t O \sim{k} O$ and $\t O$ is monotone.  Then $k(fp(\t O)) = fp(O).$
\end{proposition}
\begin{proof} Let $\t L, \t O, L$ and $O$ be as above.   Let $\t c$ be the central element of $k^{-1}(x)$.   Once again, because $\t O$ is monotone and $\t c$ is a central element, for each ordinal $\alpha$, the elements $(O^\beta(\t c))_{\beta < \alpha}$ form a chain.  As such, there must be an ordinal $\gamma$ for which $\t O^\gamma(\t c)$ is a fixpoint of $\t O$.   Therefore, it suffices to show that for each ordinal $\alpha$, $\t O^\alpha(\t c) \in k^{-1}(x)$.  The fact that  $k^{-1}(x)$ is closed under application of $O$ takes care of the case where $\alpha$ is a successor ordinal.  If $\alpha$ is a limit ordinal, then $\t O^\alpha(\t c) \in k^{-1}(x)$ because $k$ is chain continuous.
\end{proof}

Because each pair of comparable elements of $\t L$ form a chain, chain continuity of $k$ implies that $k$ is also order-preserving.  Therefore, we have the following result concerning the least fixpoints of $\t O$ and $O$.
 
\begin{proposition} \label{lfpcorr} Let  $\langle \t L, \leq' \rangle $ and $\langle L, \leq \rangle$ be complete lattices, such that $\t L \lsim{k} L$.  Let $\t O$ be a monotone operator on $\t L$ and $O$ a monotone operator on $L$, such that  $\t O \sim{k} O$.  Then $k(lfp(\t O)) = lfp(O)$ .
\end{proposition}

So far, we have shown that if a monotone operator $\t O$ $k$-mimics $O$, the fixpoints and the least fixpoint of $O$ can be found by simply applying $k$ to each fixpoint or, respectively, the least fixpoint of $\t O$.  Of course, this means that if an operator $O$ can be mimicked by a {\em stratifiable} operator $\t O$, we can incrementally construct the fixpoints and the least fixpoint of $O$, by means of the components of $\t O$.

If the operator $O$ is an approximation, however, we are also often also interested in its stable and well-founded fixpoints.   The following proposition allows us to treat these in a similar way.

\begin{proposition} \label{othermimics} Let $\t L, L$ be complete lattices, such that  $\t L \lsim{k} L$.  Let $\sk$ be the function from $\t L^2$ to $L^2$ which maps each pair $(\t x,\t y)$ to $(k(\t x), k(\t y))$.  Let $\t A$ be an  approximation on $\t L^2$ and $A$ an  approximation on $L^2$, such that $\t A \sim{\sk} A$.\\
 Then $\lowsop{\t A} \sim{k} \lowsop{A}$, $\upsop{\t A} \sim{k} \upsop{A}$ and $ \C_{\t A} \sim{\sk} \C_{A}$.  %Moreover, if $\t A$ and $A$ are exact and $\t O$ and $O$ are the unique operators they approximate, then $\t O \sim{k} O$.
\end{proposition}
\begin{proof} Let $\t L, L, \t A, A, \sk, k$ be as above. We will show that $\lowsop{\t A} \sim{k} \lowsop{A}$.  The proof of $\upsop{\t A} \sim{k} \upsop{A}$ is similar.  Because $\lowsop{A}(L^2) \subseteq A^1(L^2)$, $\lowsop{A}(L^2) \subseteq k(\t L)$.   By definition, for each $\t y \in \t L$, $k(\lowsop{\t A}(\t y)) = k(lfp(\t A^1(\cdot, \t y)))$, which because of theorem \ref{mimicsfp} equals $lfp(k(\t A^1(\cdot, \t y)) = lfp(A^1(\cdot, k(\t y)))$.
\end{proof}

As each partial stable operator $\C_A$ is by definition monotone, this proposition shows that if an approximation $A$ is $\sk$-mimicked by an approximation $\t A$, its stable and well-founded fixpoints can be found by simply applying $\sk$ to the stable fixpoints or, respectively, the well-founded fixpoint of $\t A$.  Moreover, if $\t A$ and $A$ are exact and the unique operators approximated by $\t A$ and $A$ are  monotone, the fixpoints and least fixpoint of the operator approximated by $A$ can be found by applying $k$ to the fixpoints or, respectively, the least fixpoint of the operator approximated by $\t A$.  

Once again, this means that if an approximation $A$ can be $\sk$-mimicked by a  stratifiable approximation $\t A$, we can incrementally construct the stable and well-founded fixpoints of $A$, by means of the components of $\t A$.  Similarly, if an exact approximation $A$ of a monotone operator $O$ can be $\sk$-mimicked by a stratifiable exact approximation $\t A$ of a monotone operator $\t O$, we can incrementally construct the fixpoints and least fixpoint of $O$, by means of the components of $\t A$.

\section{Applications}
\label{apps}

The general, algebraic framework of stratifiable operators developed in the previous section, allows us to easily and uniformly prove splitting theorems for logics with a fixpoint semantics.  We will demonstrate this, by applying the previous results to logic programming (Section \ref{lp}), auto-epistemic logic (Section \ref{ael}) and default logic (Section \ref{dl}).

 As noted earlier, for each of these formalisms there exists a class of approximations, such that the various kinds of fixpoints of the approximation $A_T$ associated with a theory $T$ correspond to the models of $T$ under various semantics for this formalism.

In the case of logic programming, these approximations operate on a (lattice isomorphic to) product lattice.  Therefore, it is possible to derive splitting results for logic programming by the following method: First, we need to identify syntactical conditions, such that for each logic program $P$ satisfying these conditions, the corresponding approximation $A_P$ is stratifiable.  Our results then show that the fixpoints, least fixpoint, stable fixpoints and well-founded fixpoint of such an approximation $A_P$ (which correspond to the models of $P$ under various semantics for logic programming) can be incrementally constructed from the components of $A_P$.  Of course, in order for this result to be of any practical use, one also needs to be able to actually construct these components.  Therefore, we will also present a procedure of deriving new programs $P'$ from the original program $P$, such that the approximations associated with these programs $P'$ are precisely those components.

For auto-epistemic logic and default logic the situation is, however, slightly more complicated, because the approximations which  define the semantics of these formalism do not operate on a product lattice.  Therefore, in these cases, we cannot simply follow the above procedure.  Instead, we  need to preform the additional step of first finding approximations $\t A_T$ on a product lattice ``similar'' to the original lattice, which ``mimic'' the original approximations $A_T$ of theories $T$.  The results of this section then show that the various kinds of fixpoints of $A_T$ can be found from the corresponding fixpoints of $\t A_T$.  Therefore, we can split theories $T$ by stratifying these new approximations $\t A_T$ and, as these $\t A_T$ {\em are} operators on a product lattice, this can be done by the above procedure.  In other words, we then just need to determine syntactical conditions which suffice to ensure that $\t A_T$ is stratifiable and present a way of constructing new theories $T'$ from $T$, such that the components of $\t A_T$ correspond to approximations associated with these new theories.

\subsection{Logic Programming}
\label{lp}

\subsubsection{Syntax and semantics}
\label{lpsands}
% This presentation is based on \cite{DMT00a}. 

For simplicity, we will only deal with propositional logic programs.  Let $\Sigma$ be an alphabet, i.e.~a collection of  symbols which are called {\em atoms}.  A {\em literal} is either an atom $p$ or the negation $\lnot q$ of an atom $q$.  A logic program is a set of {\em clauses} of the following form:
\begin{equation*}
h \leftarrow b_1, \ldots, b_n.
\end{equation*}
Here, $h$ is a atom and the $b_i$ are literals.  For such a clause $r$, we denote by $head(r)$ the atom $h$ and by $body(r)$ the set $\{b_1,\ldots, b_n\}$ of literals.

Logic programs can be interpreted in the lattice $\langle 2^\Sigma, \subseteq \rangle$, i.e.~the powerset of $\Sigma$.  This set of {\em interpretations} of $\Sigma$ is denoted by $\I_\Sigma$.  Following the framework of approximation theory, we will, however,  interpret programs in the bilattice $B_\Sigma = \I_\Sigma^2$.  In keeping with the intuitions presented in section \ref{approxintro}, for such a pair $(X,Y)$, the interpretation $X$ can be seen as representing an {\em under}estimate of the set of true atoms, while $Y$ represents an {\em over}estimate.  Or, to put it another way, $X$ contains all atoms which are {\em certainly} true, while $Y$ contains atoms which are {\em possibly} true.  These intuitions lead naturally to the following definition of  the truth value of a propositional formula.

\begin{definition} Let $\phi, \psi$ be  propositional formula in an alphabet $\Sigma$, $a$ an atom of $\Sigma$ and let $(X,Y) \in B_\Sigma$.  We define
\begin{itemize}
\item $H_{(X,Y)}(a) = {\bf t}$ if $a \in X$; otherwise,$ H_{(X,Y)}(a) = {\bf f}$;
\item $H_{(X,Y)}(\phi \land \psi) = {\bf t}$ if $H_{(X,Y)}(\phi) = {\bf t}$ and $H_{(X,Y)}(\psi) = {\bf t}$; otherwise, $H_{(X,Y)}(\phi \land \psi) = {\bf f}$;
\item $H_{(X,Y)}(\phi \lor \psi) = {\bf t}$ if $H_{(X,Y)}(\phi) = {\bf t}$ or $H_{(X,Y)}(\psi) = {\bf t}$; otherwise, $H_{(X,Y)}(\phi \lor \psi) = {\bf f}$;
\item $H_{(X,Y)}(\lnot \phi) = {\bf t}$ if $H_{(Y,X)}(\phi) = {\bf f}$; otherwise, $H_{(X,Y)}(\lnot \phi) = {\bf f}$.
\end{itemize}
\end{definition}

Note that to evaluate the negation of a formula $\lnot \phi$ in a pair $(X,Y)$, we actually evaluate $\phi$ in $(Y, X)$.  Indeed, the negation of a formula will be certain if the formula itself is not possible and vice versa.
Using this definition, we can now define the following operator on $B_\Sigma$.

\begin{definition}  Let $P$ be a logic program with an alphabet $\Sigma$.  The operator $\T_P$ on $B_\Sigma$ is defined as:
\begin{gather*}
\T_P(X,Y) = (U_P(X,Y), U_P(Y,X)),
\end{gather*}
with
%\begin{multline*}
$U_P(X,Y) = \{ p \in \Sigma \mid \exists r \in P: head(r) = p, H_{(X,Y)}(body(r)) = {\bf t} \}.$
%\end{multline*}
\end{definition}

When restricted to consistent pairs of interpretation, this operator $\T_P$ is the well known 3-valued Fitting operator \cite{fitting}.  In \citeN{DMT00a},  $\T_P$ is shown to be a symmetric approximation.  Furthermore, it can be used to define most of the ``popular'' semantics for logic programs: the operator which maps an interpretation $X$ to $U_P(X,X)$ is the well known (two-valued) $T_P$-operator \cite{lloyd}; the partial stable operator of $\T_P$ is the Gelfond-Lifschitz operator $\curly{GL}$ \cite{vGRS91}.  Fixpoints of $T_P$ are {\em supported models} of $P$, the least fixpoint of $\T_P$ is the Kripke-Kleene model of $P$, fixpoints of $\curly{GL}$ are (four-valued) stable models of $P$ and its least fixpoint is the well-founded model of $P$.

\subsubsection{Stratification}
\label{lpstrat}

Our discussion of the stratification of logic programs will be based on the dependecies between atoms, which are expressed by a logic program.  These induce the following partial order on the alphabet of the program. 

\begin{definition}  Let $P$ be a logic program with alphabet $\Sigma$.  The {\em dependency order} $\leq_{dep}$ on $\Sigma$ is defined as: for all $p,q \in \Sigma$:
\begin{equation*}
p \leq_{dep} q \qquad iff \qquad \exists r \in P: q = head(r), p \in body(r).
\end{equation*}
\end{definition}

To illustrate this definition, consider the following small program:
\begin{equation*}
E = \left\{
\begin{aligned}
p \leftarrow & \lnot q, \lnot r.\\
q \leftarrow &\lnot p, \lnot r.\\
s \leftarrow & p, q.
\end{aligned}
\right\}.
\end{equation*}

The dependency order of this program can be graphically represented by the following picture:
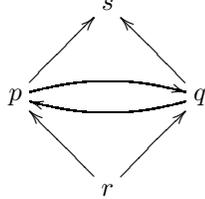
\xymatrix{
&s\\
p \ar[ur] \ar@/^/[rr] &&q \ar[ul] \ar@/^/[ll]\\
&r \ar[ur] \ar[ul]
}

In other words, $r \leq_{dep} p, r \leq_{dep} q, p \leq_{dep} q, q \leq_{dep} p, s \leq_{dep} p$ and $s \leq_{dep} q$.

Based on this dependency order, the concept of a {\em splitting} of the alphabet of a logic program can be defined.

\begin{definition} \label{splitting}  Let $P$ be a logic program with alphabet $\Sigma$.  A {\em splitting} of $P$ is a partition $(\Sigma_i)_{i \in I}$ of $\Sigma$, such that the well-founded order $\preceq$ on $I$ agrees with the dependency order $\leq_{dep}$ of $P$, i.e.~if $p \leq_{dep} q$, $p \in \Sigma_i$ and $q \in \Sigma_j$, then $i \preceq j$.
\end{definition}

For instance, the following partition is a splitting of the program $E$:  $\Sigma_0 = \{r\}$, $\Sigma_1 = \{p,q\}$ and $\Sigma_2 = \{s\}$ (with the index set $I$ being the totally ordered set $\{0,1,2\}$).  

If $(\Sigma_i)_{i \in I} $ is a partition of a logic program $P$ with alphabet $\Sigma$, the product lattice $\otimes_{i \in I} 2^{\Sigma_i}$ is clearly isomorphic to the powerset $2^\Sigma$.  We can therefore view the operator $\T_P$ of such a program as being an operator on the bilattice of this product lattice, instead of on the original lattice $B_\Sigma$.  Moreover, if such a partition is a splitting of a logic program $P$, the $\T_P$-operator on this product lattice is stratifiable.

\begin{theorem}  Let $P$ be a logic program and let $(\Sigma_i)_{i \in I}$ be a splitting of this program.  Then the operator $\T_P$ on the bilattice of the product lattice $\displaystyle \bigotimes_{i \in I} 2^{\Sigma_i}$  is stratifiable.
\end{theorem}

\begin{proof} Let $\Sigma_j \in S$ and $(X,Y), (X', Y') \in B_\Sigma$, such that $\res{X}{\preceq i} = \res{X'}{\preceq i}$ and $\res{Y}{\preceq i} = \res{Y'}{\preceq i}$.  It suffices to show that for each clause with an atom from $\Sigma_j$ in its head, $\pt_{(X,Y)}(body(c)) = \pt_{(X',Y')}(body(c))$.  By definition \ref{splitting}, this is trivially so.
\end{proof}

By theorem \ref{approxmain}, this theorem implies that, for a stratifiable program $P$, it is possible to stratify the operators $T_P, \T_P$ and $\curly{GL}$.  In other words, it is possible to split logic programs w.r.t.~the supported model, Kripke-Kleene, stable model and well-founded semantics.  Moreover, the supported, Kripke-Kleene, stable and well-founded models of $P$ can be computed from, respectively, the  supported, Kripke-Kleene, stable and well-founded models of the components of the operator $\T_P$. 

In order to be able to perform this construction in practice, however, we also need a more precise characterization of these components.  We will now show how to construct new logic programs from the original program, such that these components correspond to an operator associated with these new programs.  First, we will define the restriction of a program to a subset of its alphabet.

\begin{definition}  Let $P$ be a logic program with a splitting $(\Sigma_i)_{i \in I}$.  For each $i \in I$, the program $P_i$ consists of all clauses which have an atom from $\Sigma_i$ in their head.
\end{definition}

In the case of our example, the program $E$ is partitioned in $\{E_0, E_1, E_2\}$ with $E_0 = \{\}$, $E_1 = \{p \leftarrow \lnot q, \lnot r.\ q \leftarrow \lnot p, \lnot r.\}$ and $E_2 = \{s \leftarrow p, q.\}$.

If $P$ has a splitting $(\Sigma_i)_{i \in I}$, then clearly such a program $P_i$ contains, by definition, only atoms from $\bigcup_{j \preceq i} \Sigma_j$.  When given a pair $(U,V)$ of interpretations of $\bigcup_{j \prec i} \Sigma_j$, we can therefore construct a program containing only atoms from $\Sigma_i$ by replacing each other atom by its truth-value according to $(U,V)$.

\begin{definition}  Let $P$ be a logic program with a splitting $(\Sigma_i)_{i \in I}$.  For each $i \in I$ and $(U,V) \in \res{B_\Sigma}{\prec i}$, we define $\lval{P_i}{(U,V)}$ as the new logic program $P'$, which results from replacing each literal $l$ whose atom is in $\bigcup_{j \prec i} \Sigma_j$ by $H_{(U,V)}(\{l\})$.
\end{definition}

Of course, one can further simplify such a program by removing all clauses containing {\bf f} and just omitting all atoms {\bf t}.  Programs constructed in this way are now precisely those which characterize the components of the operator $\T_P$. 

\begin{theorem}  Let $P$ be a logic program with a splitting $(\Sigma_i)_{i \in I} $.  For each $i \in I$, $(U,V) \in \res{B_\Sigma}{ \prec i}$ and $(A,B) \in B_{\Sigma_i}$:
\begin{equation*} 
(\T_P)_{i}^{(U,V)}(A,B) = (U_{\lval{P_i}{(U,V)}}(A,B),U_{\lval{P_i}{(V,U)}}(B,A)).
\end{equation*}
\end{theorem}
\begin{proof} Let $i, U, V, A$ and $B$ be as above. Then because the order $\preceq$ on $I$ agrees with the dependency order of $P$,  $(\T_P)_{i}^{(U,V)}(A,B) = (A', B')$, with 
\begin{align*}
A' = &\{p \in \Sigma_i \mid \exists r \in P: head(r) = p, H_{(U \ext A ,V \ext B)}(body(r)) = {\bf t} \};\\
B' = &\{p \in \Sigma_i \mid \exists r \in P: head(r) = p, H_{(V \ext B ,U \ext A)}(body(r)) = {\bf t} \}.
\end{align*}
We will show that $A' = \T_{\lval{P_i}{(U,V)}}(A,B)$; the proof that $B' = \T_{\lval{P_i}{(V,U)}}(B,A)$ is  similar.  Let $r$ be a clause of $P$, such that $head(r) \in \Sigma_i$.   Then $H_{(U \ext A, V \ext B)}(body(r)) = {\bf t}$ iff $H_{(U,V)}(l) = {\bf t}$ for each literal $l$ with an atom from $\bigcup_{j \prec i} \Sigma_j)$ and $H_{(A,B)}(l') = {\bf t}$ for each literal $'l$ with an atom from $\Sigma_i$.  Because for each literal $l$ with an atom from $\bigcup_{j \prec i} \Sigma_j)$,  $H_{(U,V)}(l) = {\bf t}$ precisely iff $l$ was replaced by {\bf t} in $\lval{P_i}{(U,V)}$, this is in turn equivalent to $H_{(A, B)}(\lval{r}{(U,V)}) = {\bf t}$ (by $\lval{r}{(U,V)}$ we denote the clause which replaces $r$ in $\lval{P_i}{(U,V)}$). 
\end{proof} 

It is worth noting that this theorem implies that a component $(\T_P)_i^{(U,V)}$ is, in contrast to the operator $\T_P$ itself, not necessarily exact.

With this final theorem, we have all which is needed to incrementally compute the various fixpoints of the operator $\T_P$.  We will illustrate this process by computing 
the well-founded model of our example program $E$.  %The Kripke-Kleene fixpoint of $E_0$ is $(\{\}, \{\})$.   Replacing the atom $r$ in $E_1$ by its truth-value according to this interpretation, yields the new program $\lval{E_1}{(\{\},\{\})} = \{p \leftarrow q, q \leftarrow p\}$.  The Kripke-Kleene fixpoint of this program is $(\{\}, \{p, q\})$.   Replacing the atoms $p$ and $q$ in $E_2$ by their truth-value according to the interpretation$(\{\}, \{p, q\})$, gives the program $E_2' = \lval{E_2}{(\{\}, \{p, q\})} = \{\}$.  Similarly, the pair of interpretations $(\{p, q\}, \{\})$ yields the program $ E_2'' = \lval{E_2}{(\{p, q\}, \{\})} =  \{s\}$.  The least fixpoint of the operator $(U_{E_2'}, U_{E_2''})$ is $(\{\}, \{s\})$.  Therefore, the Kripke-Kleene fixpoint of the entire program $E$ is $(\{\} \cup \{\} \cup \{ \}, \{\} \cup \{p, q\} \cup \{s\}) = (\{\}, \{p,q,s\})$.
%The well-founded fixpoint of  $E$ can be derived in a similar manner.  
Recall that this program  is partitioned into the programs $E_0 = \{\}$, $E_1 = \{p \leftarrow \lnot q, \lnot r.\ q \leftarrow \lnot p, \lnot r.\}$ and $E_2 = \{s \leftarrow p,q.\}$.
The well-founded model of $E_0$ is $(\{\}, \{\})$.  Replacing the atom  $r$ in $E_1$ by its truth-value according to this interpretation, yields the new program $\lval{E_1}{(\{\},\{\})} = \{p \leftarrow \lnot q.\ q \leftarrow \lnot p.\}$.  The well-founded model of this program  is $(\{\},\{p, q\})$.  Replacing the atoms $p$ and $q$ in $E_2$ by their truth-value according to the pair of interpretations $(\{\},\{p, q\})$, gives the new program $E_2' = \lval{E_2}{(\{\},\{p, q\})} = \{\}$.  Replacing these by their truth-value according to the pair of interpretations $(\{p,q\},\{\})$, gives the new program $E_2'' =\lval{E_2}{(\{p, q\},\{\})} = \{s\}$. The well-founded fixpoint of $(U_{E_2'}, U_{E_2''})$ is $(\{\}, \{s\})$.   Therefore, the well-founded model of the entire program $E$ is $(\{\} \cup \{\} \cup \{\}, \{\} \cup \{p, q\} \cup \{s\}) = (\{\}, \{p,q,s\})$.

Of course, it also possible to apply these results to more complicated programs.  Consider for instance the following program in the natural numbers:
\begin{equation*}
Even = 
\left\{ \begin{aligned} even(0)&.\\
odd(X + 1)& \leftarrow even(X).\\
even(X + 1)& \leftarrow odd(X).
\end{aligned}\right\}
\end{equation*}
which can be seen as an abbreviation of the infitine propositional logic program:\\
\begin{align*}
even(0)&.\\
odd(1)& \leftarrow even(0).\\
even(1)& \leftarrow odd(0).\\
odd(2)& \leftarrow even(1).\\
even(2)& \leftarrow odd(1).\\
& \vdots 
\end{align*}
Clearly, the operator $\T_{Even}$ is stratifiable w.r.t.~to the partition\begin{equation*} (\{even(n), odd(n)\})_{n \in \mathbb{N}}\end{equation*} (using the standard order on the natural numbers) of the alphabet $\{even(n) \mid n \in \mathbb{N}\} \cup \{odd(n) \mid n \in \mathbb{N}\}$.  The component  $(\T_{Even})_0$ of this operator corresponds to the program $Even_0 = \{even(0).\}$, which has $\{even(0)\}$ as its only fixpoint.  Let $n \in \mathbb{N}$ and $U_n = \{even(i) \mid i < n, i \text{ is even}\} \cup \{odd(i) \mid i < n, i \text{ is odd}\}$.  Clearly, if $n$ is even, the component $(T_{Even})_n^{(U_n,U_n)}$ corresponds to the program $\{even(n).\}$, while if $n$ is odd $(T_{Even})_n^{(U_n,U_n)}$ corresponds to the program $\{odd(n).\}$.  This proves that the supported, Kripke-Kleene, stable and well-founded models of the program $Even$ all contain precisely those atoms $even(n)$ for which $n$ is an even natural number and those  atoms $odd(n)$ for which $n$ is an odd natural number.

\subsubsection{Related work}

\citeN{LT94} proved a splitting theorem for
logic programs under the stable model semantics; similar results were
 indepently obtained by \citeN{eiter97}. In one
respect, these results extend ours, in the sense that they apply to
logic programs with an extended syntax, also called {\em answer set
programs}, in which disjunction in the head of clauses is allowed and
in which two kinds of negation (negation-as-failure and classical
negation) can be used. While our results could easily be extended to incorporate the two negations, the
extension to disjunction in the head is less straightforward. However,
the fact that the stable model semantics for disjunctive logic
programs can also be characterized as a fixpoint semantics
\cite{leone95}, seems to suggest that our approach could be used to
obtain similar results for this extended syntax as well.  In future
work, we plan to investigate this further.  Current work into
extending approximation theory to also capture the semantics of this kind of
programs \cite{pelov04}, makes this possibility seem even more
interesting.

In another respect, our results are more general than those of
Lifschitz and Turner.  When considering only programs in the syntax
described here, our results generalize their results to include the
supported model, Kripke-Kleene and well-founded semantics as
well. This makes our results also applicable to extensions of logic
programming which, unlike answer set programming, are not based on the stable model semantics. One example in the context of deductive databases is
datalog which is based on well-founded semantics. Another example is the formalism of
ID-logic \cite{Denecker2000d,denecker04}.  This is an extension of classical logic with non-monotonic inductive definitions, the semantics of which is given by the well-founded semantics. Our stratifiability results give insight in
modularity and compositionality aspects in both logics.

While the results discussed above are, as far as we know, the ones most similar to ours, there are a number of other works which should be also be mentioned. 
Extending work from \citeN{sofiemarc}, \citeN{denecker04} recently discussed a number of modularity results for the aforementioned ID-logic.  The main difference with our work, is that these results are aimed at supporting syntactical transformations on a {\em predicate} level, whereas we have considered propositional progams.

In order to further motivate and explain the well-founded model semantics, \citeN{przymusinski98} defined the {\em dynamic stratification} of a program.  The level of an atom in this stratificiation is based on the number of iterations it takes the Gelfond-Lifschitz operator to determine the truth-value of this atom.\footnote{To be a bit more precise, $p$ belongs to level $\Sigma_i$ iff $i$ is the minimal $j$ for which $\mathcal{GL}\uparrow j = (I, J)$ and either $p \in I$ or $p \not \in J$.}  As such, this stratification precisely mimics the computation of the well-founded model and is, therefore, the tightest possible stratification of a program under the well-founded semantics.  However, as there exist no syntactic criteria which can be used to determine whether a certain stratification is the dynamic stratification of a program -- in fact, the only way of deciding this is by actually constructing the well-founded model of the program -- this concept cannot be used to perform the kind of static, upfront splitting which is our goal.

\subsection{Auto-epistemic logic}
\label{ael}

In this section, we will first describe the syntax of auto-epistemic logic and  give a brief overview, based on \citeN{DMT03}, of how a number of different semantics for this logic can be defined using concepts from approximation theory.  Then, we will follow the previously outlined methodology to prove concrete splitting results for this logic.

\subsubsection{Syntax and Semantics}
\label{aelsands}

Let $\L$ be the language of propositional logic based on a set of atoms $\Sigma$.  Extending this language with a modal operator $K$, gives a language $\L_K$ of modal propositional logic.  An auto-epistemic theory is a set of fomulas in this language $\L_K$.  For such a formula $\phi$, the subset of $\Sigma$ containing all atoms which appear in $\phi$, is denoted by $At(\phi)$; atoms which appear in $\phi$ at least once outside the scope of the model operator $K$ are called {\em objective} atoms of $\phi$ and the set of all objective atoms of $\phi$ is denoted by $At_O(\phi)$.  A {\em modal subformula} is a formula of the form $K(\psi)$, with $\psi$ a formula.

To illustrate, consider the following example:
\begin{equation*}
F = \{   \phi_1 = p \lor \lnot K p\ ;\ \
  \phi_2 = K(p \lor q) \lor q
   \}
\end{equation*}.
The objective atoms $At_O(\phi_2)$ of $\phi_2$ are $\{q\}$, while the atoms $At(\phi_2)$ are $\{p, q\}$.  The formula $K(p \lor q)$ is a modal subformula of $\phi_2$.

An {\em interpretation} is a subset of the alphabet $\Sigma$.  The set of all interpretations of $\Sigma$ is denoted by $\I_\Sigma$, i.e.~$\I_\Sigma = 2^\Sigma$.  A {\em possible world structure} is a set of interpretations, i.e.~the set of all possible world structures $\W_\Sigma$ is defined as $2^{\I_\Sigma}$.  Intuitively, a possible world structure  sums up all ``situations'' which are possible.  It therefore makes sense to order these according to inverse set inclusion to get a {\em knowledge order} $\leq_k$, i.e.~for two possible world stuctures  $Q,Q'$, $Q \leq_k Q'$ iff $Q \supseteq Q'$.  Indeed, if a possible world structure contains {\em more} possibilities, it actually contains {\em less} knowledge. 
%The set of all possible world structures for an alphabet $\Sigma$ is denoted by $\W_\Sigma$.

In case of the example, the following picture shows a part of the lattice $\W_{At(F)}$:

\def\objectstyle{\scriptstyle}
\xymatrix{ 
&&\{\} \\
\{\{\}\} \ar[urr] & \{\{p\}\} \ar[ur] &&   \{\{q\}\}\ar[ul]  &  \{\{p,q\}\} \ar[ull]\\
\cdots & \cdots & \cdots & \cdots & \cdots\\
\{\{\},\{p\}, \{q\}\} & \{\{\},\{q\}, \{p, q\}\} && \{\{\},\{p\}, \{p, q\}\} & \{\{p\}, \{q\},\{p,q\}\} \\
&& \ar[ul] \ar[ull] \ar[ur] \ar[urr] \I_{\{p,q\}} 
}

Following \citeN{DMT03}, we will define the semantics of an auto-epistemic theory by an operator on the bilattice $\B = \W_\Sigma^2$.  An element $(P,S)$ of $\B$ is known as a {\em belief pair} and is called {\em consistent} iff $P \leq_k S$.  %Intuitively, if $(P,S)$ is a consistent belief pair, we can view $P$ as representing beliefs according to a conservative  view, while $S$ can be regarded as a representation of a liberal view.  The following function evaluates the truth of a formula according to these intuitions.
In a consistent belief pair $(P,S)$, $P$ can be viewed as describing what must {\em certainly} be known, i.e.~as giving an {\em under}estimate of what is known, while $S$ can be viewed as denoting what might {\em possibly} be known, i.e.~as giving an {\em over}estimate.   Based on this intuition, there are two ways of estimating the truth of modal formulas according to $(P,S)$: we can either be {\em conservative}, i.e.~assume a formula is false unless we are sure it must be true, or we can be {\em liberal}, i.e.~assume it is true unless we are sure it must be false.  To conservatively estimate the truth of a formula $K \phi$ according to $(P,S)$, we simply have to check whether $\phi$ is surely known, i.e.~whether $\phi$ is known in the underestimate $P$.  To conservatively estimate the truth of a formula $\lnot K \phi$, on the other hand, we need to determine whether $ \phi$ is definitely unknown; this will be the case if $\phi$ cannot possibly be known, i.e.~if $\phi$ is not known in the overestimate $S$.  The following definition extends these intuitions to reach a conservative estimate of the truth of arbitrary formulas.  Note that  the objective atoms of such a formula are simply interpreted by an interpretation $X \in \I_\Sigma$.

\begin{definition} For each $(P,S) \in \B, X \in \I_\Sigma$, $a \in \Sigma$ and formulas $\phi, \phi_1$ and $\phi_2$, we inductively define $\H_{(P,S),X}$ as:
\begin{itemize}
\item $\H_{(P,S), X}(a) = {\bf t}$ iff $a \in X$ for each atom $a$; 
\item $\H_{(P,S),X}(\varphi_1 \land \varphi_2) = \true$, iff $\H_{(P,S),X}(\varphi_1)=\true$ and $\H_{(P,S),X}(\varphi_2)=\true$; 
\item $\H_{(P,S),X}(\varphi_1 \lor \varphi_2) = \true$, iff $\H_{(P,S),X}(\varphi_1)=\true$ or $\H_{(P,S),X}(\varphi_2)=\true$;
\item $\H_{(P,S),X}(\lnot\varphi)= \lnot \H_{(S,P),X}(\varphi)$;
\item $\H_{(P,S),X}(K\varphi)=\true$ iff $\H_{(P,S),Y}(\varphi)=\true$ for all $Y \in P$.
\end{itemize}
\end{definition}

%For a consistent belief pair $(P,S)$ and a formula $\phi$, $\H_{(P,S), X}(\phi)$ represents an {\em underestimate} for the truth of $\phi$, while $\H_{(S,P), X}(\phi)$ is an {\em overestimate}.  In particular, for each $P \leq_k Q \leq_k S$, $\H_{(P,S), X}(\phi) \leq \H_{(Q,Q), X}(\phi) \leq \H_{(S,P), X}(\phi)$ (with ${\bf f} \leq {\bf t}$). 
 It is worth noting that an evaluation $\H_{(Q,Q), X}(\phi)$, i.e.~one in which all that might possibly be known is also surely known, corresponds to the standard $S_5$ evaluation \cite{ch95epistemic}.  
%This also offers an intuitive explanation for why  $\H_{(P,S),X}(\lnot \phi)$ is defined as $\lnot \H_{(S,P),X}(\phi)$: in order to overestimate what we {\em do} know, we need to underestimate what we {\em don't} know and vice versa.  
Note also that the evaluation  $\H_{(P,S),X}(K \phi)$ of a modal subformula $K \phi$ depends only on $(P,S)$ and not on $X$.  We sometimes emphasize this by writing $\H_{(P,S),\cdot}(K \phi)$.  Similarly, $\H_{(P,S),X}(\phi)$ of an objective formula $\phi$ depends only on $X$ and we sometimes write $\H_{(\cdot,\cdot),X}(\phi)$ . %, it suffices to know $(P,S)$, while to evaluate an objective formula, it suffices to know $X$

As mentioned above, it is also possible to liberally estimate the truth of modal subformulas. Intuitively, we can do this by assuming that everything which might be known, {\em is} in fact known and that everything which might be unknown, i.e.~which is not surely known, is in fact unknown.  As such, to liberally estimate the truth of a formula according to a pair $(P,S)$, it suffices to treat $S$ as though it were describing what we surely know and $P$ as though it were describing what we might know.  In others words, it suffices to simply switch the roles of $P$ and $S$, i.e.~$\H_{(S,P),\cdot}$ provides a liberal estimate of the truth of modal formulas. 

These two ways of evaluating formulas can be used to derive a new, more precise belief pair $(P',S')$ from an original pair $(P,S)$.  First, we will focus on constructing the new overestimate $S'$. As $S'$ needs to overestimate knowledge, it needs to contain as few interpretations as possible.  This means that $S'$ should consist of only those interpretations, which manage to satisfy the theory even if the truth of its modal subformulas is underestimated.  So, $S' = \{ X \in \I_\Sigma \mid \forall \phi \in T: \H_{(P,S), X}(\phi) = {\bf t}\}$.  Conversly, to construct the new underestimate $P'$, we need as many interpretations as possible.  This means that $P'$ should contain all interpretations which satisfy the theory, when liberally evaluating its modal subformulas.  So, $P' = \{ X \in \I_\Sigma \mid\forall \phi \in T: \H_{(S,P), X}(\phi) = {\bf t}\}$.
These intuitions motivate the following definition of the operator $\D_T$ on $\B$:
\begin{equation*}
\D_T(P,S) = (\dtp{T}(S,P), \dtp{T}(P,S)),
\end{equation*}
with $\dtp{T}(P,S) = \{ X\in \I_\Sigma \mid \forall \phi \in T: \H_{(P,S), X}(\phi)={\bf t}\}$.

This operator is an approximation \cite{DMT03}.  Moreover, since \begin{equation*} \D_T^1(P,S) = \dtp{T}(S,P) = \D_T^2(S,P),
								 \end{equation*} and \begin{equation*} \D_T^2(P,S) = \dtp{T}(P,S) = \D_T^1(S,P),\end{equation*} it is by definition symmetric and therefore approximates a unique operator on $\W_\Sigma$, namely the operator $D_T$ \cite{Moo84}, which maps each $Q$ to $\dtp{T}(Q,Q)$.  As shown in \citeN{DMT03}, these operators define a family of  semantics for a theory $T$:
\begin{itemize}
\item fixpoints of $D_T$ are {\em expansions} of $T$ \cite{Moo84},
\item fixpoints of $\D_T$ are {\em partial expansions} of $T$ \cite{16946}, 
\item the least fixpoint of $\D_T$ is the {\em Kripke-Kleene fixpoint} of $T$ \cite{16946}, 
\item fixpoints of $\lowsop{\D_T}$ are {\em extensions} \cite{DMT03} of $T$, 
\item fixpoints of $\C_{\D_T}$ are {\em partial extensions} \cite{DMT03} of $T$, and 
\item the least fixpoint of  $\C_{\D_T}$ is the {\em well-founded model} of $T$  \cite{DMT03}.
\end{itemize}
These various dialects of auto-epistemic logic differ in their treatment of ``ungrounded'' expansions \cite{Kon87}, i.e.~expansions which arize from cyclicities such as $Kp \rightarrow p$. 

When calculating the models of a theory, it is often useful to split the calculation of $\dtp{T}(P,S)$ into two separate steps: In a first step, we evaluate each modal subformula of $T$ according to $(P,S)$ and in a second step we then compute all models of the resulting propositional theory.  To formalize this, we introduce the following notation: for each formula $\phi$ and $(P,S) \in \B$, the formula $\lval{\phi}{P,S}$ is inductively defined as:
\begin{itemize}
\item $\lval{a}{P,S} = a$ for each atom $a$; 
\item $\lval{(\varphi_1 \land \varphi_2)}{P,S} = \lval{\varphi_1}{P,S} \land  \lval{\varphi_2}{P,S}$;
\item $\lval{(\varphi_1 \lor \varphi_2)}{P,S} = \lval{\varphi_1}{P,S} \lor  \lval{\varphi_2}{P,S}$;
\item $\lval{(\lnot\varphi)}{P,S} = \lnot (\lval{\varphi}{S,P})$;
\item $\lval{(K\varphi)}{P,S} = \H_{(P,S),\cdot}(K\varphi)$. 
\end{itemize}
%, i.e.~we replace each $K\phi$ appearing in the scope of an {\em even} number of negations by $\H_{(P,S),\cdot}(K \phi)$ and each $K\phi$ appearing in the scope of an {\em odd} number of negations by $\H_{(S,P),\cdot}(K \phi)$.  The resulting new theory $T'$ is denoted by $\lval{T}{P,S}$.  
For a theory $T$, we denote $\{\lval{\phi}{P,S} \mid \phi \in T\}$ by $\lval{T}{P,S}$.
Because clearly 
\begin{equation*}\H_{(\cdot,\cdot),X}(\lval{T}{P,S}) = {\bf t}\text{  iff  }\H_{(P,S),X}(T) = {\bf t}, 
\end{equation*}
it is the case that for each $(P,S) \in \B$:
\begin{equation*}\D_T^u(P,S) = \{X \in \I_\Sigma \mid \H_{(\cdot, \cdot), X}(\lval{T}{P,S})\}.\end{equation*}
%it then suffices to --- in a second step --- calculate all models of the (propositional) theory $\lval{T}{P,S}$.

To illustrate, we will construct the Kripke-Kleene model of our example theory $F = \{p \lor \lnot K p;\ K(p \lor q) \lor q\}$.  This computation starts at the least precise element $(\I_{\{p,q\}}, \{\})$ of $\curly{B}_{\{p, q\}}$.  We first construct the new underestimate $\dtp{F}(\{\}, \I_{\{p,q\}})$.   It is easy to see that 
\begin{equation*} \H_{(\{\}, \I_{\{p,q\}}), \cdot}(\lnot Kp) = \lnot  \H_{(\I_{\{p,q\}}, \{\}), \cdot}(Kp) = \lnot {\bf f} = {\bf t},
\end{equation*}
and 
\begin{equation*}
\H_{(\{\},\I_{\{p, q\}}), \cdot}(K(p \lor q)) = {\bf t}.
\end{equation*}
 Therefore, $\lval{F}{\{\},\I_{\{p,q\}}} = \{p \lor {\bf t};\ q \lor {\bf t}\}$ and $\dtp{F}(\{\}, \I_{\{p,q\}}) =  \I_{\{p, q\}}. $
 Now, to compute the new overestimate $\dtp{F}(\I_{\{p,q\}}, \{\})$, we note that
\begin{equation*}
\H_{(\I_{\{p,q\}}, \{\}), \cdot}(\lnot Kp) = \lnot \H_{(\{\},\I_{\{p,q\}}), \cdot}(Kp) = \lnot {\bf t} = {\bf f},
\end{equation*}
and
\begin{equation*}
\H_{(\I_{\{p,q\}}, \cdot), \cdot}( K(p \lor q)) = {\bf f}.
\end{equation*}
Therefore, $\lval{F}{(\I_{\{p,q\}}, \{\})} = \{p \lor {\bf f};\ q \lor {\bf f}\}$ and
$\dtp{F}(\I_{\{p,q\},\{\}}) =  \{\{p, q\}\}.$
So, $\D_T(\I_{\{p,q\}}, \{\}) =  (\I_{\{p,q\}}, \{\{p, q\}\})$.  

To compute $\dtp{F}(\{p, q\}, \I_{\{p,q\}})$, we note that it is still the case that
\begin{equation*}
\H_{(\cdot, \I_{\{p,q\}}), \cdot}(\lnot Kp) = \H_{(\{\{p, q\}\},\cdot), \cdot}(K(p \lor q) = {\bf t}.
\end{equation*}
So, $\dtp{F}(\{\{p, q\}\}, \I_{\{p,q\}}) = \I_{\{p, q\}}$.  Similary, 
\begin{equation*}
\H_{(\cdot, \{\{p, q\}\}), \cdot}(\lnot K p) = \H_{(\I_{\{p, q\}}, \cdot), \cdot}(\lnot K p) = {\bf f}. 
\end{equation*}
So, $\dtp{F}(\I_{\{p,q\}},\{\{p, q\}\}) = \{\{p, q\}\}$.  Therefore, $(\I_{\{p,q\}}, \{\{p, q\}\})$ is the least fixpoint of $\D_F$, i.e.~the Kripke-Kleene model of $F$.

\subsubsection{Stratification}
\label{aelstrat}

Let $(\Sigma_i)_{i \in I}$ be a partition of the alphabet $\Sigma$, with $\langle I, \preceq \rangle$ a well-founded index set.  For an interpretation $X \in \I_\Sigma$, we denote  the intersection $X \cap \Sigma_i$ by $\res{X}{\Sigma_i}$.  For a possible world structure $Q$,  $\{\res{X}{\Sigma_i} \mid X \in Q \}$ is denoted by $\res{Q}{\Sigma_i}$.

In the previous section, we defined the semantics of auto-epistemic logic in terms of an operator on the bilattice $\B = \W_\Sigma^2$.  However, for our purpose of stratifying auto-epistemic theories, we are interested in the bilattice $\Bt$ of the product lattice $\Wt = \bigotimes_{i \in I} \W_{\Sigma_i}$. 
An element of this product lattice consists of a number of possible interpretations for each level $\Sigma_i$.  As such, if we choose for each $\Sigma_i$ one of its interpretations, the union of these ``chosen'' interpretations interprets the entire alphabet $\Sigma$.  Therefore, the set of all possible ways of choosing one interpretation for each $\Sigma_i$, determines a set of possible interpretations for $\Sigma$, i.e.~an element of $\W_\Sigma$.  More formally, we define:
\begin{align*}
  \kappa : \Wt  \rightarrow \W_\Sigma : \tilde Q  \mapsto \{\bigcup_{i \in I} S(i) \mid S \in \otimes_{i \in I} \tilde Q(i)\}.
\end{align*}
Similary, $\Bt$ can be mapped to $\B$ by the function $\kb$, which maps each $(\t P, \t S)  \in \Bt$ to $(\kappa(\t P), \kappa(\t S))$.

This function $\kappa$ is, however, not an isomorphism.  Indeed, unlike $\W_\Sigma$, elements of $\Wt$ cannot express that an interpretation for a level $\Sigma_i$ is possible in combination with a {\em certain} interpretation for another level $\Sigma_j$, but {\em not} with a different interpretation for $\Sigma_j$.  For instance, if we split the alphabet $\{p, q\}$ of our example $F$ into $\Sigma_0 = \{p\}$ and $\Sigma_1 = \{q\}$, the element $\{\{p, q\}, \{\}\}$ of $\W_\Sigma$ is not in $\kappa(\Wt)$, because it expresses that $\{p\}$ is only a possible interpretation for $\Sigma_0$ when $\Sigma_1$ is interpreted by $\{q\}$ and not when $\Sigma_1$ is interpreted by $\{\}$.  To make this more precise, we introduce the following concept.

\begin{definition} A possible world structure $Q \in \W_\Sigma$ is {\em disconnected} w.r.t.~a partition $(\Sigma_i)_{i \in I}$ of its alphabet iff for all possible worlds $X, Y \in Q$ and for each $i \in I$, $(\res{Y}{\Sigma_i} \cup \bigcup_{j \neq i} \res{X}{\Sigma_j} ) \in Q$.
\end{definition}

\begin{proposition}\label{charkappa}
\begin{equation*}
\kappa(\Wt) = \{ Q \in \W_\Sigma \mid Q \text{ is disconnected}\}.
\end{equation*}
\end{proposition}
\begin{proof}
We will first prove the inclusion from left to right.  Let $\Qt$ be an element of $\Wt$ and let $X,Y$ be elements of $\kappa(\Qt)$.  Then, by the definition of $\kappa$, for each $i \in I$, there are elements $A_i, B_i$ in $\Qt(i)$, such that $X  = \cup_{i \in I} A_i$ and $Y =  \cup_{i \in I} B_i$.  Clearly, then,  $(B_i \cup \bigcup_{j \neq i} A_i)$ is also in $\kappa(\tilde Q)$.  Therefore, $\kappa(\t Q)$ is disconnected.

To prove the other direction, let $Q$ be a disconnected element of $\W_\Sigma$.  Let $\tilde Q$ be the following element of $\Wt$:
\begin{equation*}
 \tilde Q : I \rightarrow \bigcup_{i \in I} \W_{\Sigma_i}: i \mapsto \{ A_i \in \I_{\Sigma_i} \mid \exists X \in Q:  A_i = \res{X}{\Sigma_i} \}
\end{equation*}
Then \begin{multline*}
\kappa(\tilde Q) = \displaystyle \big\{\bigcup_{i \in I} A_i \mid \forall i \in I: A_i \in \t Q(i) \big\} =\big\{ ( \bigcup_{i \in I} \res{X_i}{\Sigma_i}) \mid \forall i \in I: X_i \in Q \big\} = Q.
     \end{multline*}
\end{proof}

\newcommand{\lfp}{\text{\emph{lfp}}}
\newcommand{\nat}{\mathbb{N}}           

We now restrict our attention to a class of theories whose models are all disconnected. 

\begin{definition}\label{defstratael}
 An auto-epistemic theory $T$ is {\em stratifiable} w.r.t.~a partition  $(\Sigma_i)_{i \in I}$ of its alphabet and a well-founded order $\preceq$ on $I$, if there exists a partition $\{T_i\}_{i \in I}$ of $T$ such that for each $i\in I$ and  $\phi \in T_i$: $At_O(\phi) \subseteq \Sigma_i$ and
 $At(\phi) \subseteq \displaystyle \bigcup_{j \preceq i} \Sigma_j$.
\end{definition}

To illustrate, our example theory $F$ is stratifiable w.r.t.~the partition $\Sigma_0 = \{p\}$, $\Sigma_1 = \{q\}$ of its alphabet $\{p, q\}$ and the corresponding partition of $F$  is $F_0 = \{ p \lor \lnot Kp \}$, $F_1 = \{ K(p \lor q) \lor q\}$. 

Clearly, for a stratifiable theory, the evaluation $\H_{(P,S),X}(\phi)$ of a formula $\phi \in T_i$ only depends on the value of $(P,S)$ in strata $j \preceq i$ and that of $X$ in stratum $i$.

\begin{proposition}\label{HeqH} \saet Let $i \in I$ and $\phi \in T_i$.  Then for each $(P,S)$, $(P', S') \in \B$ and $X,X' \in \I_\Sigma$, such that $\res{X}{\Sigma_i} = \res{X'}{\Sigma_i}$ and $\res{P}{\cup_{j \preceq i}\Sigma_j} = \res{P'}{\cup_{j \preceq i}\Sigma_j}$ and $\res{S}{\cup_{j \preceq i}\Sigma_j} = \res{S'}{\cup_{j \preceq i}\Sigma_j}$ , $\H_{(P,S),X}(\phi) =  \H_{(P',S'),X'}(\phi)$.
\end{proposition}

This proposition can now be used to show that only disconnected belief pairs are relevant for the operator $\D_T$.

\begin{proposition} \saet \label{charim} Then each $\D_T(P, S)$ is disconnected.
\end{proposition}
\begin{proof}
Let $(P,S) \in \B$ and $X,Y \in \dtp{T}(P,S)$.  By proposition \ref{HeqH}, for each $Z \in \I_\Sigma$, such that, for some $i \in I$, $\res{Z}{\Sigma_i} = \res{X}{\Sigma_i}$ and, $\forall j \neq i$, $\res{Z}{\Sigma_j} = \res{Y}{\Sigma_j}$, $Z \in \dtp{T}(P,S)$.  Therefore $\dtp{T}(P, S)$ and $\D_T(P,S)$ are both disconnected.
\end{proof}

This result suggests that the fact that $\kappa$ is not surjective should not pose any problems, because we can simply forget about possible world structures of $\W_\Sigma$ that do not correspond to elements of our product lattice $\Wt$.  However, there is another difference between the lattices $\Wt$ and $\W_\Sigma$, that needs to be taken into account.  Indeed, besides not being surjective, $\kappa$ is also not injective.  Concretely, $\Wt$ contains multiple ``copies'' of the empty set, that is, for any $\t Q \in \Wt$, as soon as for some $i \in I$, $\t Q(i) = \{\}$, it is the case that $\kappa(\t Q) = \{\}$.  

Let us introduce some notation and terminology.  We call an possible world structure $Q \in \W_\Sigma$ {\em consistent} if $Q \neq \{\}$; the set of all consistent $Q$ is denoted by $\W_\Sigma^c$. A belief pair $(P,S)$ is called consistent if both $P$ and $S$ are consistent; the set of all consistent belief pairs is denoted $\B^c$.  Similarly, $\t Q \in \Bt$ is called consistent if $\kappa(\t Q) \neq \{\}$ and the set of all consistent $\t Q$ is denoted as $\Wt^c$.  Finally, a belief pair $(\t P, \t S)$ is called consistent if both $\kappa(\t P) \neq \{\}$ and $\kappa(\t S) \neq \{\}$ and we denote the set of all consistent pairs $(\t P, \t S)$ as $\Bt^c$. 

We will often need to eliminate inconsistent possible world structures from our considerations.  Intuitively, the reason for this is that, when constructing a stratification, we need every stratum $i$ to be completely independent of all strata $j$ for which $j \succ i$.  However, if an inconsistency occurs at level $j$, then this could affect the way in which a lower level $i$ is interpreted, because it will eliminate {\em all} possible worlds.  Mathematically, this problem manifest itself by the fact that the equality $\res{\kappa(\t Q)}{\preceq i} = \kappa(\res{\t Q}{\preceq i})$ only holds for consistent possible world structures $\t Q$.

We now summarize some obvious properties of $\kappa$.

\begin{proposition} \label{th:kappaprop} The function $\kappa$ has the following properties.
\begin{enumerate}
\item $\kappa$ is order preserving;
\item $\kappa$ is an embedding of $\Wt^c$ into $\W_\Sigma^c$ and an isomorphism between $\Wt^c$ and the set of all disconnected possible world structures in $\W_\Sigma^c$;
\item For all consistent $\t Q$, $\kappa(\res{\t Q}{\preceq i}) = \res{\kappa(\t Q)}{\preceq i}$.
\end{enumerate}
\end{proposition}

Because of the differences between the lattices $\Bt$ and $\B$ outlined above, we cannot directly stratify the operator $\D_T$.  Instead, we will define an intermediate operator $\Dt$ on $\Bt$, which is stratifiable by construction and whose consistent fixpoints are related to consistent fixpoints of $\D_T$.  We define this operator $\Dt$ in such a way that, for any belief pair $(\t P, \t S) \in \Bt$, the $i$th level of $\Dt(\t P, \t S)$ will be constructed using only the theory $T_i$ and the restriction $\res{(\t P, \t S)}{\preceq i}$ of $(\t P, \t S)$.

\begin{definition} \saet Let $(\t P, \t S)$ be in $\Bt$.  We define $\Dt^u(\t P, \t S) = \t Q$, with for each $i \in I$:
\begin{equation*}\t Q(i) =
 \{X \in \I_{\Sigma_i} \mid \forall \phi \in T_i: \H_{\kb(\res{\t P}{\preceq i}, \res{\t S}{\preceq i}), X}(\phi) = {\bf t}\}.
\end{equation*}
Furthermore, $\Dt(\t P, \t S) = (\Dt^u(\t S, \t P),\Dt^u(\t P, \t S))$ and $\t D_T(\t Q) = \Dt^u(\t Q, \t Q)$.
\end{definition}

Observe that we could equivalently define the $i$th level of $\Dt^u(\t P, \t S)$ as the set $Mod(\lval{T_i}{\kb(\res{\t P}{\preceq i}, \res{\t S}{\preceq i})})$ of  models of the propositional theory $\lval{T_i}{\kb(\res{\t P}{\preceq i}, \res{\t S}{\preceq i})}$.

Let us now first show that, like its counterpart $\D_T$, this operator is also an approximation.

\begin{proposition} \saet \label{Dtapp} Then $\Dt$ is an approximation.
\end{proposition}

\begin{proof}  Let $(\t P, \t S), (\t P', \t S') \in \Bt$, such that $(\tilde P,\tilde S) \leq_p (\tilde P',\tilde S')$.   Because then $(\t S, \t P) \geq (\t S', \t P')$, we only need to show that $\D_T^u(\t P, \t S) \geq_\otimes \D_T^u(\t P', \t S')$.  Let $i \in I$.  Because $\kappa$ is  order-preserving, $\kb(\res{\tilde P}{\preceq i},\res{\tilde S}{\preceq i}) \leq_p \kb(\res{\tilde P'}{\preceq i},\res{\tilde S'}{\preceq i})$.  From \citeN{DMT03}, we know  this implies that for each  $\phi$ of $T_i$ and $X$ in $\I_{\Sigma_i}$, if $\H_{\kb(\res{\tilde P}{\preceq i},\res{\tilde S}{\preceq i}), X}(\phi) = {\bf t}$ then $\H_{\kb(\res{\tilde P'}{\preceq i},\res{\tilde S'}{\preceq i}), X}(\phi) = {\bf t}$.  Hence, $\Dt^u(\tilde P, \tilde S)(i) \subseteq \dtp{T}(\tilde P', \tilde S')(i)$. 
\end{proof}

We can relate the consistent fixpoints of $\Dt$ to those of $\D_T$, using the following result.  

\begin{proposition} For all consistent $(\t P, \t S) \in \Bt^c$, $\kb(\t \D_T(\t P, \t S)) = \D_T(\kb(\t P, \t S))$.  \label{th:mimics}
\end{proposition}
\begin{proof}
Let $(\t P, \t S) \in \Bt^c$.  By symmetry of the operators $\Dt$ and $\D_T$, it suffices to show that $\kappa(\t \D_T^u(\t P, \t S)) = \D_T^u(\kb(\t P, \t S))$.  Because for $i \neq j$, the objective atoms of $T_i$ and $T_j$ are disjoint, it is a trivial property of propositional logic that $Mod(\lval{T}{\kb(\t P,\t S)}$ consists precisely of all worlds of the form $\cup_i X_i$ for which $X_i \in Mod(\lval{T_i}{\kappa(\t P), \kappa(\t S)})$.  Now, let $\t Q$ be the element of $\Bt$ that maps every $i \in I$ to $Mod(\lval{T_i}{\kappa(\t P), \kappa(\t S)})$.  We then have that $\kappa(\t Q) = \D_T^u(\kappa(\t P), \kappa(\t S))$.  It therefore suffices to show that $\t Q = \t \D_T^u(\t P, \t S)$, that is, that for all $i\in I$, $\lval{T_i}{\kappa(\t P), \kappa(\t S)} = \lval{T_i}{\kappa(\res{\t P}{\preceq i}), \kappa(\res{\t S}{\preceq i})}$.  Because $T_i$ contains only modal literals in the alphabet $\cup_{j \preceq i} \Sigma_j$, we already have that  $\lval{T_i}{\kappa(\t P), \kappa(\t S)} =  \lval{T_i}{\res{\kappa({\t P})}{\preceq i}, \res{\kappa(\t S)}{\preceq i}}$.  Because $(\t P, \t S)$ is consistent, we also have that $(\res{\kappa(\t P)}{\preceq i}, \res{\kappa(\t S)}{\preceq i}) = (\kappa(\res{\t P}{\preceq i}), \kappa(\res{\t S}{\preceq i}))$, which proves the result.
\end{proof}

We already know that $\kappa$ is an isomorphism between $\Bt^c$ and the set of all consistent, disconnected belief pairs in $\B$, and that all consistent fixpoints of $\D_T$ are also disconnected.  Therefore, the above Proposition \ref{th:mimics} now directly implies the following result.

\begin{proposition}
The set of all $\kappa(\t P, \t S)$ for which $(\t P, \t S)$ is a consistent fixpoint of $\Dt$ is equal to the set of all consistent fixpoints of $\D_T$.
\label{th:fpcorr}
\end{proposition}

A similar correspondence also holds for the consistent stable fixpoints of these two operators.  Our proof of this depends on the following result.

\begin{proposition}  For all consistent $\t S \in \Wt^c$, $\kappa(C^\downarrow_{\t \D_T}(\t S)) = C^\downarrow_{\D_T}(\kappa(\t S))$.\label{th:sop}
\end{proposition}
\begin{proof}
Recall that the operator $C^\downarrow_{\D_T}$ is defined as mapping each $S$ to $\lfp(\D_T^u(\cdot, S))$ and, similarly, $C^\downarrow_{\Dt}$ maps each $\t S$ to $\lfp(\Dt^u(\cdot, \t S))$.  Let $\t S \in \Wt^c$ and $S = \kappa(\t S)$.  The values $C^\downarrow_{\D_T}(S)$ and  $C^\downarrow_{\Dt}(\t S)$ can be constructed as the limit of, respectively, the ascending sequences $(Q_n)_{0 \leq n}$ and $(\t Q_n)_{0 \leq n}$, defined as follows: $Q_0$ is the bottom element $\I_\Sigma$ of $\W_\Sigma$ and for every $n > 0$, $Q_n =  \D_T^u(Q_{n-1}, S)$; similarly, $\t Q_0$ is the bottom element of $\Wt$, that is, for all $i \in I$, $\t Q_0(i) = \I_{\Sigma_i}$, and for all $n > 0$, $\t Q_n = \Dt^u(\t Q_{n - 1}, \t S)$.  It  suffices to show that, for all $n \in \nat$, $Q_n = \kappa(\t Q_n)$.  

We prove this by induction over $n$.
For the base case, it is clear that $\kappa(\t Q_0) = Q_0$.  Now, suppose that the equality holds for $n$.  Then $\t Q_{n + 1} = \t \D_T^u(\t Q_n, \t S)$, which we must prove equal to $Q_{n + 1} = \D_T^u(Q_n, S)$.  By the induction hypothesis, this last expression is equal to $\D_T^u(\kappa(\t Q_n), \kappa(\t S))$.  We distinguish two cases.  First, let us assume that $\t Q_{n}$ is consistent.  By Proposition \ref{th:mimics}, it is then the case that $\kappa(\t \D_T^u(\t Q_n, \t S)) = \D_T^u(\kb(\t Q_n, \t S))$, which is precisely what needs to be proven.  Second, assume that $\kappa(\t Q_n) = Q_n = \{\}$.  Because $\t Q_{n +1} \geq_k \t Q_n$ and $\kappa$ is order-preserving, $\kappa(\t Q_{n+1}) = \{\}$.  Moreover, because also $Q_{n + 1} \geq_k Q_n = \{\}$, we have that $Q_{n+ 1} = \{\}$, which means that, here too, we get the desired equality $\kappa(\t Q_{n + 1}) = Q_{n+1}$.
\end{proof}

Together with the fact that $\kappa$ is an isomorphism between $\Wt^c$ and the set of all consistent, disconnected possible world structures, this result now directly implies the following correspondence between consistent stable fixpoints.

\begin{proposition}
The set of all $\kappa(\t P, \t S)$ for which $(\t P, \t S)$ is a consistent stable fixpoint of $\Dt$ is equal to the set of all consistent stable fixpoints of $\Dt$.
\label{th:sfpcorr}
\end{proposition}

We can now summarize the content of Propositions \ref{th:fpcorr} and \ref{th:sfpcorr} in the following theorem.

\begin{theorem} \label{th:notperma}
Let $T$ be a stratifiable theory.   The set of all $\kb({\t P},{\t S})$ for which  $(\t P, \t S)$ is a consistent fixpoint (consistent stable fixpoint, respectively) of $\Dt$ is equal to the set of all $(P, S)$, for which $(P,S)$ is a consistent fixpoint (consistent stable fixpoint) of $\D_{T}$.
\end{theorem}

We now define a class of theories, for which (partial) expansions and (partial) extensions cannot be inconsistent.

\begin{definition} An auto-epistemic theory $T$ is {\em permaconsistent} if every propositional theory $T'$ that can be constructed from $T$ by replacing all non-nested occurrences of modal literals by $\bf t$ or $\bf f$ is consistent.
\end{definition}

Observe that, contrary to what the above definition might suggest, we do not actually need  to check {\em every} assignment in order to determine whether a theory is permaconsistent.  Indeed, it suffices to only consider the worst case assignment, in which every positive occurrence of a modal literal is replaced by {\bf f} and every negative occurrence is replaced by {\bf t}.  Moreover, we can also  check permaconsistency for every stratum separately, because, for a stratifiable theory $T$, $T$ is permaconsistent iff for every $i \in I$, $T_i$ is permaconsistent.

Clearly, for a permaconsistent theory $T$, every belief pair in the image of $\D_T$ or $\Dt$ is consistent and, therefore, all fixpoints or stable fixpoints of these operators must be consistent as well.  As such, Theorem \ref{th:notperma} implies that the fixpoints and stable fixpoints of $\D_T$ and $\t \D_T$ coincide, which of course implies that also the least fixpoint and well-founded fixpoint of these two operators coincide.  In summary, we obtain the following result.

\begin{theorem}\label{th:perma}
Let $T$ be a stratifiable and permaconsistent theory.   The set of all $\kb({\t P},{\t S})$ for which $(\t P, \t S)$ is a fixpoint, the Kripke-Kleene fixpoint, a stable fixpoint, or the well-founded fixpoint of $\Dt$ is equal to the set consisting of, respectively, all fixpoints, the Kripke-Kleene fixpoint, all stable fixpoints, or the well-founded fixpoint of $\D_{T}$.
\end{theorem}

This property does not hold for theories that are not permaconsistent.  We illustrate this by the following example.  Let $T$ be the theory $\{ Kp \Leftarrow p; \lnot K(p \lor q) \}$.  This $T$ is stratifiable with respect to the partition $\Sigma_0 = \{p\}$, $\Sigma_1 = \{q\}$, with the corresponding partition of $T$ being $T_0 = \{  Kp \Leftarrow p \}$ and $T_1 = \{ \lnot K(p \lor q) \}$.  However, $T$ is not permaconsistent, because, for instance, $\{{\bf t} \Leftarrow p; \lnot {\bf t}\}$ is not consistent.  The operator $\D_T$ now has as its Kripke-Kleene fixpoint the pair $(\{ \{p,q\} \}, \{\})$, as can be seen from the following computation:
\begin{center}
\begin{tabular}{r@{=\ (}c@{,\ }c@{)}}
$(\bot_k, \top_k)$ & $\I_{\{p,q\}}$ & $\{\}$\\
$\D_T(\bot_k, \top_k)$ & $Mod(\lval{T}{\{\}, \I_{\{p,q\}}})$ & $Mod(\lval{T}{\I_{\{p,q\}}, \{\}})$ \\
& $Mod({\bf t} \Leftarrow p; \lnot {\bf f})$ & $Mod({\bf f} \Leftarrow p; \lnot {\bf t})$ \\
& $\I_{\{p,q\}}$ & $\{\}$ \\
\end{tabular}
\end{center}
The Kripke-Kleene fixpoint of $\Dt$ can be constructed as follows:
\begin{center}
\begin{tabular}{r@{}l@{=\ (}c@{,\ }c@{)}}
$(\t \bot, \t \top)$ & (0) & $\I_{\{p\}}$ & $\{\}$\\
& (1)  & $\I_{\{q\}}$ & $\{\}$\\
$\Dt(\t \bot, \t \top)$ & (0) & $Mod(\lval{T_0}{\{\}, \I_{\{p\}}})$ & $Mod(\lval{T_0}{\I_{\{p\}}, \{\}})$ \\
& & $Mod({\bf t} \Leftarrow p)$ &$Mod({\bf f} \Leftarrow p)$ \\
& & $\I_{\{p\}}$ & $\{\{\}\}$\\
& (1) &  $Mod(\lval{T_1}{\{\}, \I_{\{p,q\}}})$ &  $Mod(\lval{T_1}{\I_{\{p,q\}}, \{\}})$ \\
& & $Mod(\lnot {\bf f})$ & $Mod(\lnot {\bf t})$\\
& & $\I_{q}$ & $\{\}$\\
$\Dt^2(\t \bot, \t \top)$ & (0) &  $Mod(\lval{T_0}{\{\{\}\}, \I_{\{p\}}})$ &$Mod(\lval{T_0}{\I_{\{p\}}, \{\{\}\}})$ \\
& & $Mod({\bf f} \Leftarrow p)$ &$Mod({\bf f} \Leftarrow p)$ \\
& & $\{\{\}\}$ & $\{\{\}\}$\\
& (1) &  $Mod(\lval{T_1}{\{\}, \I_{p,q}})$ &  $Mod(\lval{T_1}{\I_{\{p,q\}}, \{\}})$\\
& & $\I_{\{q\}}$ & $\{\}$\\
$\Dt^3(\t \bot, \t \top)$ & (0) &  $Mod(\lval{T_0}{\{\{\}\}, \{\{\}\}})$ &$Mod(\lval{T_0}{\{\{\}\}, \{\{\}\}})$ \\
& & $\{\{\}\}$ & $\{\{\}\}$\\
& (1) &  $Mod(\lval{T_1}{\{\}, \{\{\},\{q\}\}})$ &  $Mod(\lval{T_1}{\{\{\},\{q\}\}, \{\}})$\\
& & $\I_{\{q\}}$ & $\{\}$\\
\end{tabular}
\end{center}
So, we find that applying $\kappa$ to the Kripke-Kleene fixpoint of $\Dt$ yields $(\{\{\},\{q\}\}, \{\})$, which is not equal to the Kripke-Kleene fixpoint $(\I_{\{p,q\}}, \{\})$ of $\D_T$.

Using the correspondences between consistent (stable) fixpoints of $\Dt$ and $\D_T$, we can now proceed to analyse $T$ by looking at $\Dt$ and moreover, because $\Dt$ is by construction stratifiable, we can do so by applying our algebraic stratification results to this operator.  Concretely, we can incrementally construct its (stable) fixpoints, using its component operators.  Of course, for this result to be useful, we also need to know what these component operators actually are.  It turns out that a component on level $i$ corresponds to a theory, that can be constructed from $T_i$, by replacing certain formulas by their truth value according to a belief pair $(\t U, \t V) \in \res{\Bt}{\prec i}$.  Before showing this for all stratifiable theories, we first consider the following, more restricted class of theories.

\begin{definition} A theory $T$ is {\em modally separated} w.r.t.~to a partition $(\Sigma_i)_{i \in I}$ of its alphabet iff there exists a corresponding partition $(T_i)_{i \in I}$ of $T$, such that for each $i \in I$ and $\phi \in T_i$
\begin{itemize}
\item $At_O(\phi) \subseteq \Sigma_i$,
\item for each modal subformula $K \psi$ of $\phi$, either $At(\psi) \subseteq \Sigma_i$ or $At(\psi) \subseteq \bigcup_{j \prec i} \Sigma_j$.
\end{itemize}
\end{definition}

Clearly, all modally separated theories are stratifiable.  The fact that each modal subformula of a level $T_i$ of a modally separated theory $T$ contains either only atoms from $\Sigma_i$ or only atoms from a strictly lower level, makes it easy to construct the components of its $\t \D_T$-operator. Replacing all modal subformulae of a level $T_i$ which contain only atoms from a strictly lower level $j \prec i$, by their truth-value according to a belief pair  $(\t U, \t V) \in \res{\B}{\prec i}$ results in a ``conservative theory'' $T^c$, while replacing these subformulae by their truth-value according to $(\t V, \t U)$  yields a ``liberal theory'' $T^l$.  The pair $(\D_{T^l}^u, \D_{T^c}^u)$ is then precisely the component $(\t \D_T)_i^{(\t U, \t V)}$ of $\t \D_T$.

To make this more precise, we inductively define the following transformation $\lval{\phi}{U,V}_i$ of a formula $\phi \in T_i$, given a belief pair  $(\t U, \t V) \in \res{\t \B}{\prec i}$:
\begin{itemize}
\item $\lval{a}{\t U,\t V}_i = a$ for each atom $a$; 
\item $\lval{(\varphi_1 \land \varphi_2)}{\t U,\t V}_i = \lval{\varphi_1}{\t U,\t V}_i \land \lval{\varphi_2}{\t U,\t V}_i$;
\item $\lval{(\varphi_1 \lor \varphi_2)}{\t U,\t V}_i = \lval{\varphi_1}{\t U,\t V}_i \lor  \lval{\varphi_2}{\t U,\t V}_i$;
\item $\lval{(\lnot\varphi)}{\t U,\t V}_i = \lnot (\lval{\varphi}{\t V,\t U}_i)$;
\item $  \lval{(K\varphi)}{\t U, \t V}_i  =  \begin{cases}\H_{\kb(\t U,\t V),\cdot}(K\varphi) &\text{if } At(\varphi) \subseteq \bigcup_{j \prec i}\Sigma_j;\\
 K(\varphi)& \text{if } At(\varphi) \subseteq \Sigma_i.
      \end{cases}
$
\end{itemize}

Note that this transformation $\lval{\phi}{\t U,\t V}_i$ is identical to the transformation $\lval{\phi}{P,S}$ defined earlier, except for the fact that in this case, we only replace modal subformulae with atoms from $\bigcup_{j \prec i} \Sigma_j$ and leave modal subformulae with atoms from $\Sigma_i$ untouched.

Let us now consider a component $(\t \D_T)_i^{(\t U, \t V)}$ of the $\Dt$-operator of a modally separated theory $T$.  Such a component maps each $(\t P_i, \t S_i) \in  \curly{\t B}_{\Sigma_i}$ to $\res{\Dt(\t P, \t S)}{i}$, where $(\t P, \t S)$ is any element of $\Bt$ that coincides with $(\t U, \t V)$ on all levels $j \prec i$ and with $(\t P_i, \t S_i)$ on level $i$.  As such, in the construction of a new belief pair $(\t P_i', \t S_i')$, this component will evaluate modal literals that appear in $T_i$ and whose atoms belong to $\bigcup_{j \prec i} \Sigma_j$, according to either $\res{\kb(\res{(\t P, \t S)}{\preceq i})}{\prec i}$ or $\res{\kb(\res{(\t S, \t P)}{\preceq i})}{\prec i}$.  Now, if $(\t P_i, \t S_i)$ is consistent, then $\res{\kb(\res{(\t P, \t S)}{\preceq i})}{\prec i} = (\t U, \t V)$ and $\res{\kb(\res{(\t S, \t P)}{\preceq i})}{\prec i} = (\t V, \t U)$.   It follows that we can characterize the behaviour of a component  $(\t \D_T)_i^{(\t U, \t V)}$ on consistent belief pairs as follows:  

\begin{proposition} Let $T$ be a modally separated theory. Let $i \in I$ and $(\t U,\t V) \in\res{\Bt}{\prec i}$. For all consistent $(\t P_i,\t S_i) \in  \curly{\tilde B}_{\Sigma_i}^c$: \label{compms}
\begin{equation*}
(\Dt)_i^{(\t U,\t V)} (\t P_i,\t S_i)  = (\D_{\lval{T_i}{\t V,\t U}_i}^u(\t S_i ,\t P_i),  \D_{\lval{T_i}{\t U,\t V}_i}^u(\t P_i,\t S_i)).
\end{equation*}
\end{proposition}

Now, all that remains to be done is to characterize the components of stratifiable theories which are not modally separated.  It turns out that for each stratifiable theory $T$, there exists a modally separated theory $T'$, which is  equivalent to $T$ with respect to evaluation in  disconnected possible world structures.
To simplify the proof of this statement, we recall that
each formula $\phi$ can be written in an equivalent form $\phi'$ such that each modal subformula of $\phi'$ is of the form $K(a_1 \lor \cdots \lor a_m)$, with each $a_i$ an objective literal.  This result is well-known for $S_5$ semantics and can --- using the same transformation ---  be shown to also hold for all semantics considered here\footnote{To show this, it suffices to show that each step of this transformation preserves the value of the evaluation $\H_{(P, S), X}(\phi)$.  For all steps corresponding to properties of (three-valued) propositional logic, this is trivial.  The step of transforming a formula $K(K(\phi))$ to $K(\phi)$ also trivially satisfies this requirement.  All that remains to be shown, therefore, is that $\H_{(P,S), \cdot}(K(\phi \land \psi)) = \H_{(P,S), \cdot}(K(\phi) \land K(\psi))$.  By definition, $\H_{(P,S), \cdot} (K(\phi \land \psi)) = {\bf t}$ iff  $\forall X \in P:  \H_{(P, S),X}(\phi) = {\bf t}$ and $\H_{(P, S),X}(\phi) = {\bf t}$, which in turn is equivalent to $\forall X \in P: \H_{(P,S), X} (K(\phi)) = {\bf t}$ and $\forall X \in P: \H_{(P,S), X} (K(\psi)) = {\bf t}$.}.  

\begin{proposition} \label{splitK}  Let $( P, S)$ be a disconnected element of $\B$.  Let $i \in I$, $b_1,\ldots,b_n$ literals with atoms from $\Sigma_i$ and $c_1,\ldots,c_m$ literals with atoms from $\bigcup_{j \prec i} \Sigma_j$.  Then
\begin{equation*}
\H_{(P, S), \cdot} (K(\bigvee_{j = 1..n} b_j \lor \bigvee_{j = 1..m} c_j)) = 
\H_{(P, S), \cdot} (K(\bigvee_{j = 1..n} b_j) \lor K( \bigvee_{j = 1..m} c_j)).
\end{equation*}
\end{proposition}
\begin{proof}
 By definition, \begin{equation*}
\H_{( P, S), \cdot} (K(\bigvee_{j = 1..n} b_j \lor \bigvee_{j = 1..m} c_j)) = {\bf t}
		\end{equation*}
iff 
\begin{equation*}
\forall X \in P: \H_{(\cdot, \cdot), X}(\bigvee_{j = 1..n} b_j \lor \bigvee_{j = 1..m} c_j) = {\bf t}.
\end{equation*}
 This is equivalent to 
$\forall X \in P$,  $\H_{(\cdot, \cdot), X}(\bigvee_{j = 1..n} b_j) = {\bf t}$ or $\H_{(\cdot, \cdot), X} (\bigvee_{j = 1..m} c_j) = {\bf t}$. 
Because $P$ is disconnected, it contains all possible combinations $\res{X}{\cup_{j \prec i} \Sigma_j} \cup \res{Y}{\Sigma_i} \cup \res{Z}{\cup_{j \not \preceq i} \Sigma_j}$, with $X,Y,Z \in P$.   Therefore the previous statement is in turn equivalent to for each $X,Y \in P$, $\H_{(\cdot, \cdot), \res{X}{\Sigma_i}}(\bigvee_{j = 1..n} b_j) = {\bf t}$ or $\H_{(\cdot, \cdot), \res{Y}{\cup_{j \prec i} \Sigma_j}}(\bigvee_{j = 1..m} c_j)$, which proves the result.
\end{proof}

As previously discussed, given a level $T_i$ of a stratifiable auto-epistemic theory $T$, we can  construct an equivalent theory $T_i'$ in which every modal subformula is of the form $K(a_1 \lor \cdots \lor a_n)$, with the $a_i$ objective literals.  Using the above proposition, we can further split every modal subformula of such a $T'_i$ into a part containing only symbols from $\Sigma_i$ and a part containing only symbols from $\bigcup_{j \prec i} \Sigma_j$, thus creating a modally separated theory $T_i''$, which is equivalent to $T_i$ with respect to evaluation in pairs of disconnected possible world structures.  We will denote this $T_i''$ as $\nf{T_i}$.  In the case of our example $F =\{p \lor \lnot K p;\ K(p \lor q) \lor q\}$, the modally separated theory $\nf{F} = \{p \lor \lnot K p;\ K(p) \lor K(q) \lor q\}$ is equivalent to $F$ with respect to evaluation in disconnected belief pairs.  Together with Proposition \ref{compms}, the fact that for all $(\t P, \t S) \in \Bt$ and $X \in \W_\Sigma$, $\H_{\kb(\t P, \t S), X}(T_i) = \H_{\kb(\t P, \t S), X}([T_i])$  implies the following characterization of the component operators of $\Dt$.

\begin{proposition} \label{aelcomp} 
 Let $i \in I$ and $(\t U,\t V) \in\res{\Bt}{\prec i}$. For all consistent $(\t P_i,\t S_i) \in  \curly{\tilde B}_{\Sigma_i}^c$:
\begin{equation*}
(\Dt)_i^{(\t U,\t V)} (\t P_i,\t S_i) = (\D_{\lval{\nf{T_i}}{ \t V,\t U}_i}^u(\t S_i ,\t P_i),  \D_{\lval{\nf{T_i}}{\t U,\t V}_i}^u(\t P_i,\t S_i)).
\end{equation*}
\end{proposition}

Because we already know that we can construct consistent fixpoints of the operator $\D_T$ by incrementally constructing consistent fixpoints of the component operators of $\Dt$, this result now provides the final piece of the puzzle, by showing how these component operators can be derived from the theory $T$.  For $i \in I$ and $(\t U,\t V) \in\res{\Bt}{\prec i}$, let us define that a belief pair $(\t P_i,\t S_i) \in \curly{\t B}_{\Sigma_i}$ is a {\em stratified partial expansion} of stratum $T_i$ given $(\t U,\t V)$ if it is a fixpoint of the component $(\Dt)_i^{(\t U,\t V)}$. 
A belief pair $(P, S)$ is now a consistent partial expansion of $T$ if and only if there exists a $(\t P, \t S)$, such that $\kb(\t P, \t S) = (P,S)$ and, for each $i \in I$, $(\t P, \t S)(i)$ is a consistent stratified partial expansion of $T_i$ given $\res{(\t P, \t S)}{\prec i}$.

  Note that if $\t U = \t V$, then of course $\lval{\nf{T_i}}{\t V, \t U} = \lval{\nf{T_i}}{\t U, \t V}$, which means that, on consistent belief pairs, the component $(\Dt)_i^{(\t U,\t V)}$ coincides with the operator $\t \D_{T'}$ for the theory $T' = \lval{\nf{T_i}}{\t U, \t V}_i$.  As such, the stratified partial expansions of $T_i$ given some exact pair $(\t U, \t U)$ are simply the partial expansions of the theory $\lval{\nf{T_i}}{\t U, \t U}_i$.  

Let us illustrate this be means of the example $F$, which we previously partitioned into $F_0 = \{p \lor \lnot K p \}$ and $F_1 = \{K(p \lor q) \lor q \}$.  The belief pair $(\{\{\},\{p\}\}, \{\{p\}\})$ is a consistent partial expansion of $F_0$, as can be seen from the following equations:
\begin{align*}
Mod(\lval{F_0}{\{\{p\}\},\ \{\{\},\{p\}\}}) = Mod(p \lor \lnot{\bf f}) &= \{\{\},\{p\}\};\\
Mod(\lval{F_0}{\{\{\},\{p\}\},\ \{\{p\}\}}) = Mod(p \lor \lnot{\bf t}) &= \{\{p\}\}.\\
\end{align*}
For the second level, we need to consider $\nf{F_1} = \{Kp \lor Kq \lor q\}$ and use this to construct two theories $F_1^l$ and $F_1^c$, to be used in the construction of, respectively, the underestimates and overestimates for $\Sigma_1$, that is, $F_1^l$ liberally estimates the truth of $Kp$, while $F_1^c$ estimates it conservatively:
\begin{align*}
F_1^l = \lval{\nf{F_1}}{\{\{p\}\},\ \{\{\},\{p\}\}}_1 = \{{\bf t} \lor Kq \lor q\}; \\
F_1^c = \lval{\nf{F_1}}{\{\{\},\{p\}\},\ \{\{p\}\}}_1 = \{{\bf f} \lor Kq \lor q\}.
\end{align*}
The above proposition now tells us that the component operator $(\Dt)_1^{(\{\{\},\{p\}\}, \{\{p\}\})}$ coincides with the operator $(\t \D_{F_1^l}^u, \t \D_{F_1^c}^u)$ on consistent belief pairs.  Therefore, this component has a fixpoint $(\{\{\},\{q\}\}, \{\{q\}\})$, as can be seen from the following equations:
\begin{align*}
Mod(\lval{F_1^l}{\{\{q\}\},\ \{\{\},\{q\}\}}) = Mod({\bf t} \lor {\bf t} \lor q) &= \{\{\},\{q\}\};\\
Mod(\lval{F_1^c}{\{\{\},\{q\}\}},\ \{\{q\}\}) = Mod({\bf f} \lor {\bf f} \lor q) &= \{\{q\}\}.
\end{align*}
Now, the set of all $I \cup J$ for which $I \in \{\{\}, \{p\}\}$ and  $J \in \{\{\}, \{q\}\}$ is $\I_{\{p,q\}}$, while the set of all $M \cup N$ for which $M \in \{\{p\}\}$ and $N \in \{\{q\}\}$ is the singleton $\{\{p,q\}\}$.  We therefore conclude that the belief pair $(\I_{\{p,q\}}, \{\{p,q\}\})$ is a consistent partial expansion of $T$.

So far, the above discussion has only considered partial expansions.  We can of course also look at other semantics.  Let us define a {\em stratified partial extension} of $T_i$ given $(\t U, \t V)$ as a stable fixpoint of the component $(\Dt)_i^{(\t U, \t V)}$.  We also introduce the term {\em stratified expansion} (or {\em stratified extension}) to refer to a $\t Q \in \Wt$ for which $(\t Q, \t Q)$ is a stratified partial expansion (respectively, stratified partial extension). 

\begin{theorem}  \saet  An belief pair $(P, S)$ of $\B$ is a consistent partial expansion (respectively, consistent partial extension) of $T$ iff there exists a $(\t P, \t S) \in \Bt$, such that $\kb(\t P, \t S) = (P, S)$ and for all $i \in I$, $(\t P, \t S)(i)$ is a consistent stratified partial expansion (consistent stratified partial extension) of $T_i$ given $\res{(\t P, \t S)}{\prec i}$.  A possible world structure $Q \in \W_\Sigma$ is a consistent expansion (respectively, consistent extension) of $T$ iff there exists a $\t Q \in \Wt$, such that $\kappa(\t Q) = Q$ and for all $i \in I$, $\t Q(i)$ is a consistent expansion (consistent extension) of $\lval{T_i}{\res{(\t Q, \t Q)}{\prec i}}_i$.
\end{theorem}

For permaconsistent theories, we can draw stronger conclusions.  Let us call the least stratified partial expansion (respectively, least stratified partial extension) the {\em stratified Kripke-Kleene model} ({\em stratified well-founded model}).  We then have the following result.

\begin{theorem}
\saet If $T$ is also permaconsistent, then $(P,S)$ is a partial expansion, partial extension, Kripke-Kleene model or well-founded model of $T$ iff there exists a $(\t P, \t S) \in \Bt$, such that $\kb(\t P, \t S) = (P, S)$ and for all $i \in I$, $(\t P, \t S)(i)$ is, respectively, a stratified partial expansion, stratified partial extension, stratified Kripke-Kleene model or stratified well-founded model of $T_i$ given $\res{(\t P, \t S)}{\prec i}$.  A possible world structure $Q \in \W_\Sigma$ is an expansion (extension, respectively) of $T$ iff there exists a $\t Q \in \Wt$, such that $\kappa(\t Q) = Q$ and for all $i \in I$, $\t Q(i)$ is an expansion (extension) of $\lval{T_i}{\res{(\t Q, \t Q)}{\prec i}}_i$.
\end{theorem}

\subsubsection{Related work}

In \citeN{aelsplit} and \citeN{niemela94}, it was shown that certain permaconsistent and modally separated auto-epistemic theories can be split under the semantics of expansions.  We have both extended these results to other semantics for this logic and to a larger class of theories. 

To give some intuition about the kind of theories our result can deal with, but previous work cannot, we will consider the following example (from \citeN{etherington87}):  Suppose we would like to express that we suspect a certain person of murder if we know he had a motive and if it is possible that this person is a suspect and that he is guilty. This naturally leads to following formula:
\begin{equation*}K motive \land \lnot K( \lnot suspect \lor \lnot guilty) \rightarrow suspect.
\end{equation*}
This formula is not modally separated w.r.t.~the partition
\begin{equation*}\Sigma_0 = \{guilty, motive\}, \Sigma_1 = \{suspect\}
\end{equation*}
 and, therefore, falls outside the scope of Gelfond et al.'s theorem.   Our result, however, does cover this example and allows it to be split w.r.t.~this partition.  As we will discuss in the next section on default logic, there exists an important class of default expressions, called {\em semi-normal defaults}, which typically give rize to such statements.

\subsection{Default logic}
\label{dl}

\subsubsection{Syntax and semantics}
\label{dlsands}
Let $\L$ be the language of propositional logic for an alphabet $\Sigma$.  A {\em default } $d$ is a formula
\begin{equation}\label{default}
\default{\alpha}{\beta_1,\ldots,\beta_n}{\gamma}
\end{equation}
with $\alpha,\beta_1,\ldots,\beta_n,\gamma$ formulas of $\L$.  The formula $\gamma$ is called the consequence $cons(d)$ of $d$.  A {\em default theory} is a pair $\langle D, W \rangle$, with $D$ a set of defaults and $W$ formulas of $\L$.  

\citeN{Kon87} suggested a transformation $m$ from default logic to auto-epistemic logic, which was shown by \citeN{DMT03}  to capture the semantics of default logic.  For simplicity, we will ignore the original formulation of the semantics of default logic and view this as being defined by the auto-epistemic theory $m( D,W )$.

\begin{definition} Let $\langle D,W \rangle$ be a default theory and let $d=\default{\alpha}{\beta_1,\ldots,\beta_n}{\gamma}$ be a default in $D$. Then
\begin{equation}\label{kontrans}
m(d) = (K\alpha \land \lnot K \lnot \beta_1 \land \cdots \land \lnot K \lnot \beta_n \Rightarrow \gamma)
\end{equation}
and
\begin{equation*}
m( D,W  ) = \{m(d) \mid d \in D\} \cup W.
\end{equation*}
\end{definition}

A pair $(P,S)$ of possible world structures is an expansion \cite{marek89}, a partial expansion \cite{DMT03}, an extension \cite{Rei80}, a partial extension \cite{DMT03}, the Kripke-Kleene model \cite{DMT03} or the well-founded model \cite{BS91} of a default theory $\langle D, W \rangle$ if it is, respectively, an expansion, a partial expansion, an extension, a partial extension, the Kripke-Kleene or the well-founded model of $m(D, W)$.  The semantics of extensions is the most common.

\subsubsection{Stratification}
\label{dlstrat}

We begin by defining the concept of a stratifiable default theory.

\begin{definition} \label{dlstratdef}  Let $\langle D,W \rangle$ be a default theory over an alphabet $\Sigma$.  Let $(\Sigma_i)_{i \in I}$ be a partition of $\Sigma$. $\langle D,W \rangle$ is stratifiable over this partition, if there exists a partition $\langle D_i, W_i \rangle_{i \in I}$ of $\langle D, W \rangle$ such that:
\begin{itemize}
\item For each default $d$: if an atom of $cons(d)$ is in $\Sigma_i$, then $d \in D_i$,
\item For each default $d \in D_i$: all atoms of $d$ are in $\bigcup_{j \preceq i} \Sigma_j$.
\item For each $w \in W$, if $w$ contains an atom $p \in \Sigma_i$, then $w \in W_i$.
\end{itemize}
\end{definition}

The auto-epistemic theory corresponding to a stratifiable default theory is also stratifiable (according to Definition \ref{defstratael}).

\begin{proposition} Let $\langle D,W \rangle$ be a stratifiable default theory.  Then $m( D,W )$ is a stratifiable auto-epistemic theory.
\end{proposition}

By the results of the previous section, this now immediately implies the following result.

\begin{theorem}\label{th:default}
Let  $\langle D,W \rangle$ be a stratifiable default theory.  A belief pair $(P, S)$ is a consistent partial expansion (respectively, consistent partial extension) of $\langle D, W\rangle$ iff there exists a $(\t P, \t S)$ such that $\kb(\t P, \t S) = (P, S)$ and for each $i \in I$, $(\t P, \t S)(i)$ is a consistent stratified partial expansion (consistent stratified partial extension) of the auto-epistemic theory $m( D_i, W_i)$ given $\res{(\t P, \t S)}{\prec i}$.  A possible world structure $Q$ is a consistent expansion (respectively, consistent extension) of $\langle D, W\rangle$ if there exists a $\t Q$ such that $\kappa(\t Q) = Q$ and for each $i \in I$, $\t Q(i)$ is a consistent stratified expansion (consistent stratified extension) of the auto-epistemic theory $m( D_i, W_i)$ given $\res{(\t Q, \t Q)}{\prec i}$.
\end{theorem}

Consistent stratified partial expansions and extensions of the auto-epistemic theory $T_i = m(D_i, W_i)$ can be constructed using the component theories $\lval{[T_i]}{\res{(\t S, \t P)}{\prec i}}_i$ and $\lval{[T_i]}{\res{(\t P, \t S)}{\prec i}}_i$.  Because no such component theory can contain a nested occurrence of a modal literal,  standard transformations for propositional logic can be used to bring each of its formulas into the form:
\[ \lnot (K \alpha_1 \land K \alpha_2\land\cdots\land K\alpha_m) \lor K\beta_1 \lor \cdots \lor K \beta_n \lor \gamma,\]
where the $\alpha_i$, $\beta_j$ and $\gamma$ are all propositional formulas.  Because $K \alpha_1 \land\cdots\land K\alpha_m$ is equivalent to $K (\alpha_1 \land \cdots \land \alpha_m)$, this suffices to show that each such formula can be transformed back into a default.  As such, it would be possible to reformulate the above theorem entirely in terms of default logic.

We now investigate a specific class of theories, for which it is particularly useful to restate our result in this manner.  Let us call a default theory $\langle D, W\rangle$ {\em modally separated} if the auto-epistemic theory $m(D, W)$ is modally separated or, equivalently, if for every default $d \in D_i$ of form \eqref{default}, it is the case that no formula $\phi \in \{\alpha,\beta_1,\ldots,\beta_n\}$ contains both an atom $p \in \Sigma_i$ and $q \in \bigcup_{j \prec i} \Sigma_j$.   Given a stratum $\langle D_i, W_i \rangle$ of such a theory and a belief pair $(\t U, \t V)\in \res{\Bt}{\prec i}$, we define $\lval{D_i}{\t U, \t V}_i$ as the set of defaults $d'$ that result from replacing, in every $d \in D_i$, all formulas $\phi \in \{\alpha,\beta_1,\ldots,\beta_n\}$ whose atoms belong to alphabet $\bigcup_{j \prec i} \Sigma_j$, by their truth value according to $(\t U, \t V)$.  Now, we can characterize the auto-epistemic component theory $\lval{[T_i]}{\t U, \t V}_i$ as simply being $m(\lval{D_i}{\t U, \t V}_i, W_i)$.  Therefore, Theorem \ref{th:default} implies that a possible world structure $P$ is a consistent extension of a stratifiable default theory $\langle D, W\rangle$ if and only if there exists a $\t P$ such that $\kappa(\t P) = P$ and, for each $i \in I$, $\t P(i)$ is a consistent extension of the theory $\langle \lval{D_i}{\res{(\t P, \t P)}{\prec i}}_i, W_i \rangle$.

\subsubsection{Related work}

\citeN{Tur96a} proved splitting theorems for default logic, which considered only consistent extensions.  We extend these results to the semantics of consistent partial extensions and consistent (partial) expansions.  Moreover, his results only apply to modally separated default theories.  Our results therefore not only generalize previous results to other semantics, but also to a larger class of theories.

A typical example of a default which is not modally separated but which can be split using our results, is the example from \citeN{etherington87} concerning murder suspects.  This can be formalized by the following default:
\begin{equation*}
\frac{ motive :  suspect \land guilty}{suspect}.
\end{equation*}
In the previous section, we already presented the auto-epistemic formula resulting from applying the Konolige transformation to this default and showed that it was not modally separated w.r.t.~the partition 
\begin{equation*}\Sigma_0 = \{guilty, motive\}, \Sigma_1 = \{suspect\}.\end{equation*} 
Therefore, Turner's theorem does not apply in this case, but our results do.  

Defaults such as these are typical examples of so-called {\em semi-normal defaults}, i.e.~defaults of the form 
\begin{equation*} \frac{\alpha: \beta}{\gamma} \end{equation*}
where $\beta$ implies $\gamma$.  This typically occurs because there is some formula $\delta$, such that $\beta = \gamma \land \delta$.  In such cases, the Konolige transformation will contain a formula $K(\lnot \gamma \lor \lnot \delta)$ and such defaults can therefore only be modally separated w.r.t.~stratifications in which all atoms from both $\gamma$ and $\delta$ belong to the same stratum.  Our results, however, also allows stratifications in which (all or some) atoms from $\delta$ belong to a strictly lower stratum than the atoms from $\gamma$.

\section{Conclusion}
\label{conc}

Stratification is, both theoretically and practically, an important concept in knowledge representation.  We have studied this issue at a general, algebraic level by investigating stratification of {\em operators} and {\em approximations} (Section \ref{mainsec}).  This gave us a small but very useful set of theorems, which enabled us to easily and uniformly prove splitting results for all fixpoint semantics of logic programs, auto-epistemic logic and default logic (Section \ref{apps}), thus generalizing existing results.

As such, the importance of the work presented here is threefold.  Firstly, there are the concrete, applied results of Section \ref{apps} themselves.  Secondly, there is the general, algebraic framework for the study of stratification, which can be applied to every formalism with a fixpoint semantics.  Finally, on a more abstract level, our work also offers greater insight into the principles underlying various existing stratification results, as we are able to ``look beyond'' purely syntactical properties of a certain formalism.


\begin{thebibliography}{}

\bibitem[\protect\citeauthoryear{Apt, Blair, and Walker}{Apt
  et~al\mbox{.}}{1988}]{Apt88}
{\sc Apt, K.}, {\sc Blair, H.}, {\sc and} {\sc Walker, A.} 1988.
\newblock {Towards a theory of Declarative Knowledge}.
\newblock In {\em Foundations of Deductive Databases and Logic Programming},
  {J.~Minker}, Ed. Morgan Kaufmann.

\bibitem[\protect\citeauthoryear{Baral and Subrahmanian}{Baral and
  Subrahmanian}{1991}]{BS91}
{\sc Baral, C.} {\sc and} {\sc Subrahmanian, V.} 1991.
\newblock Duality between alternative semantics of logic programs and
  nonmonotonic formalisms.
\newblock In {\em International Workshop on Logic Programming and Nonmonotonic
  Reasoning}, {A.~Nerode}, {W.~Marek}, {and} {V.~Subrahmanian}, Eds. MIT Press,
  Washington DC., 69--86.

\bibitem[\protect\citeauthoryear{Denecker}{Denecker}{2000}]{Denecker2000d}
{\sc Denecker, M.} 2000.
\newblock Extending classical logic with inductive definitions.
\newblock In {\em {1st International Conference on Computational Logic
  (CL2000)}}, {J.~{Lloyd et al.}}, Ed. Lecture Notes in Artificial
  Intelligence, vol. 1861. Springer, London, 703--717.

\bibitem[\protect\citeauthoryear{Denecker, Marek, and Truszczynski}{Denecker
  et~al\mbox{.}}{1998}]{16946}
{\sc Denecker, M.}, {\sc Marek, V.}, {\sc and} {\sc Truszczynski, M.} 1998.
\newblock {F}ixpoint 3-valued semantics for autoepistemic logic.
\newblock In {\em {P}roceedings of the 15th {N}ational {C}onference on
  {A}rtificial {I}ntelligence}. MIT Press / AAAI-Press, 840--845.

\bibitem[\protect\citeauthoryear{Denecker, Marek, and Truszczynski}{Denecker
  et~al\mbox{.}}{2000}]{DMT00a}
{\sc Denecker, M.}, {\sc Marek, V.}, {\sc and} {\sc Truszczynski, M.} 2000.
\newblock {A}pproximating operators, stable operators, well-founded fixpoints
  and applications in non-monotonic reasoning.
\newblock In {\em {L}ogic-based {A}rtificial {I}ntelligence}. The Kluwer
  International Series in Engineering and Computer Science. Kluwer Academic
  Publishers, Boston, 127--144.

\bibitem[\protect\citeauthoryear{Denecker, Marek, and Truszczynski}{Denecker
  et~al\mbox{.}}{2003}]{DMT03}
{\sc Denecker, M.}, {\sc Marek, V.}, {\sc and} {\sc Truszczynski, M.} 2003.
\newblock {U}niform semantic treatment of default and autoepistemic logics.
\newblock {\em Artificial Intelligence\/}~{\em 143,\/}~1 (Jan.), 79--122.

\bibitem[\protect\citeauthoryear{Denecker and Ternovska}{Denecker and
  Ternovska}{2004}]{denecker04}
{\sc Denecker, M.} {\sc and} {\sc Ternovska, E.} 2004.
\newblock A logic of non-monotone inductive definitions and its modularity
  properties.
\newblock In {\em {LPNMR}}. 47--60.

\bibitem[\protect\citeauthoryear{Dix}{Dix}{1995}]{dix95}
{\sc Dix, J.} 1995.
\newblock A classification theory of semantics normal logic programs: {II}.
  weak properties.
\newblock {\em Fundamenta Informaticae\/}~{\em XXII}, 257--288.

\bibitem[\protect\citeauthoryear{Eiter, Gottlob, and Mannila}{Eiter
  et~al\mbox{.}}{1997}]{eiter97}
{\sc Eiter, T.}, {\sc Gottlob, G.}, {\sc and} {\sc Mannila, H.} 1997.
\newblock Disjunctive datalog.
\newblock {\em {ACM} {T}ransactions on {D}atabase {S}ystems (TODS)\/}~{\em 22},
  364--418.

\bibitem[\protect\citeauthoryear{Erdo{\u g}an and Lifschitz}{Erdo{\u g}an and
  Lifschitz}{2004}]{erdogan04}
{\sc Erdo{\u g}an, S.} {\sc and} {\sc Lifschitz, V.} 2004.
\newblock Definitions in {A}nswer {S}et {P}rogramming.
\newblock In {\em Proc. Logic Programming and Non Monotonic Reasoning,
  LPNMR'04}. LNAI, vol. 2923. {S}pringer-{V}erlag, 185--197.

\bibitem[\protect\citeauthoryear{Etherington}{Etherington}{1988}]{etherington8%
7}
{\sc Etherington, D.} 1988.
\newblock {\em Reasoning with incomplete information}.
\newblock Research notes in {A}rtificial {I}ntelligence. Morgan Kaufmann.

\bibitem[\protect\citeauthoryear{Fitting}{Fitting}{1985}]{fitting}
{\sc Fitting, M.} 1985.
\newblock {A Kripke-Kleene Semantics for Logic Programs}.
\newblock {\em Journal of Logic Programming\/}~{\em 2,\/}~4, 295--312.

\bibitem[\protect\citeauthoryear{Fitting}{Fitting}{1991}]{fitting89}
{\sc Fitting, M.} 1991.
\newblock Bilattices and the semantics of logic programming.
\newblock {\em Journal of Logic Programming\/}~{\em 11}, 91--116.

\bibitem[\protect\citeauthoryear{Gelfond}{Gelfond}{1987}]{Gelfond87}
{\sc Gelfond, M.} 1987.
\newblock {On Stratified Autoepistemic Theories}.
\newblock In {\em Proc. of AAAI87}. Morgan Kaufman, 207--211.

\bibitem[\protect\citeauthoryear{Gelfond and Przymusinska}{Gelfond and
  Przymusinska}{1992}]{aelsplit}
{\sc Gelfond, M.} {\sc and} {\sc Przymusinska, H.} 1992.
\newblock On consistency and completeness of autoepistemic theories.
\newblock {\em Fundamenta Informaticae\/}~{\em 16,\/}~1, 59--92.

\bibitem[\protect\citeauthoryear{Ginsberg}{Ginsberg}{1988}]{gins88}
{\sc Ginsberg, M.} 1988.
\newblock Multivalued logics: a uniform approach to reasoning in artificial
  intelligence.
\newblock {\em Computational Intelligence\/}~{\em 4}, 265--316.

\bibitem[\protect\citeauthoryear{Konolige}{Konolige}{1987}]{Kon87}
{\sc Konolige, K.} 1987.
\newblock On the relation between default and autoepistemic logic.
\newblock In {\em Readings in Nonmonotonic Reasoning}, {M.~L. Ginsberg}, Ed.
  Kaufmann, Los Altos, CA, 195--226.

\bibitem[\protect\citeauthoryear{Leone, Rullo, and Scarcello}{Leone
  et~al\mbox{.}}{1995}]{leone95}
{\sc Leone, N.}, {\sc Rullo, P.}, {\sc and} {\sc Scarcello, F.} 1995.
\newblock Declarative and fixpoints characterizations of disjunctive stable
  models.
\newblock In {\em Proc. of {I}nternational {L}ogic {P}rogramming
  {S}symposium-ILPS'95}. {MIT} {P}ress, 399--413.

\bibitem[\protect\citeauthoryear{Lifschitz and Turner}{Lifschitz and
  Turner}{1994}]{LT94}
{\sc Lifschitz, V.} {\sc and} {\sc Turner, H.} 1994.
\newblock Splitting a logic program.
\newblock In {\em Proceedings of the 11th International Conference on Logic
  Programming}. MIT Press, Cambridge, MA, USA, 23--37.

\bibitem[\protect\citeauthoryear{Lloyd}{Lloyd}{1987}]{lloyd}
{\sc Lloyd, J.} 1987.
\newblock {\em Foundations of Logic Programming}.
\newblock Springer-Verlag.



\bibitem[\protect\citeauthoryear{Marek and Truszczynski}{Marek and Truszczynski}{1989}]{marek89}
{\sc Marek, V.} {\sc and} {\sc Truszczynski, M.} 2004.
\newblock Stable Semantics for Logic Programs and Default Theories.
\newblock In {\em Proceedings of the North American Conference on Logic Programming},
\newblock MIT Press, 327--334. 


\bibitem[\protect\citeauthoryear{Meyer and van der Hoek}{Meyer and van der Hoek}{1995}]{ch95epistemic}
{\sc Meyer, J.-J.Ch.} {\sc and} {\sc van der Hoek, W.} 1995.
\newblock {\em Epistemic Logic for Computer Science and Artificial Intelligence}.
\newblock Cambridge University Press.


\bibitem[\protect\citeauthoryear{Moore}{Moore}{1984}]{Moo84}
{\sc Moore, R.~C.} 1984.
\newblock Possible-world semantics for autoepistemic logic.
\newblock In {\em Proceedings of the Non-Monotonic Reasoning Workshop}. AAAI,
  Mohonk, N.Y, 344--354.



\bibitem[\protect\citeauthoryear{Pelov}{Pelov}{2004}]{pelov04}
{\sc Pelov, N.} {\sc and} {\sc Truszczynski, M.} 2004.
\newblock Semantics of disjunctive programs with monotone aggregates - an
  operator-based approach.
\newblock In {\em Proceedings of the 10th International Workshop on
  Non-Monotonic Reasoning (NMR 2004), Whistler, Canada, June 6-8, 2004,
  Proceedings}, {J.~P. Delgrande} {and} {T.~Schaub}, Eds. 327--334.






\bibitem[\protect\citeauthoryear{Niemel{\" a} and Rintanen}{Niemel{\" a} and
  Rintanen}{1994}]{niemela94}
{\sc Niemel{\" a}, I.} {\sc and} {\sc Rintanen, J.} 1994.
\newblock On the impact of stratification on the complexity of nonmonotonic
  reasoning.
\newblock {\em Journal of Applied Non-Classical Logics\/}~{\em 4,\/}~2.

\bibitem[\protect\citeauthoryear{Przymusinski}{Przymusinski}{1998}]{przymusins%
ki98}
{\sc Przymusinski, T.} 1998.
\newblock Every logic program has a natural stratification and an iterated
  least fixed point model.
\newblock In {\em Proceedings of the 8th Symposium on Principles of Database
  Systems (PODS)}. 11--21.

\bibitem[\protect\citeauthoryear{Reiter}{Reiter}{1980}]{Rei80}
{\sc Reiter, R.} 1980.
\newblock A logic for default reasoning.
\newblock {\em {A}rtificial {I}ntelligence\/}~{\em 13,\/}~1--2, 81--132.

\bibitem[\protect\citeauthoryear{Tarski}{Tarski}{1955}]{tar55}
{\sc Tarski, A.} 1955.
\newblock Lattice-theoretic fixpoint theorem and its applications.
\newblock {\em Pacific Journal of Mathematics\/}~{\em 5}, 285--309.


\bibitem[\protect\citeauthoryear{Turner}{Turner}{1996}]{Tur96a}
{\sc Turner, H.} 1996.
\newblock Splitting a default theory.
\newblock In {\em Proceedings of the 13th National Conference on Artificial
  Intelligence and the Eighth Innovative Applications of Artificial
  Intelligence Conference}. AAAI Press, 645--651.

\bibitem[\protect\citeauthoryear{{Van Gelder}, Ross, and Schlipf}{{Van Gelder}
  et~al\mbox{.}}{1991}]{vGRS91}
{\sc {Van Gelder}, A.}, {\sc Ross, K.}, {\sc and} {\sc Schlipf, J.} 1991.
\newblock {The Well-Founded Semantics for General Logic Programs}.
\newblock {\em Journal of the ACM\/}~{\em 38,\/}~3, 620--650.

\bibitem[\protect\citeauthoryear{Vennekens, Gilis, and Denecker}{Vennekens
  et~al\mbox{.}}{2004a}]{nmr}
{\sc Vennekens, J.}, {\sc Gilis, D.}, {\sc and} {\sc Denecker, M.} 2004a.
\newblock {S}plitting an operator: {A}n algebraic modularity result and its
  application to auto-epistemic logic.
\newblock In {\em {P}roceedings of {I}nternational {W}orkshop on
  {N}on-{M}onotonic {R}easoning, {W}histler, {B}ritish {C}olumbia, {C}anada}.

\bibitem[\protect\citeauthoryear{Vennekens, Gilis, and Denecker}{Vennekens
  et~al\mbox{.}}{2004b}]{iclp}
{\sc Vennekens, J.}, {\sc Gilis, D.}, {\sc and} {\sc Denecker, M.} 2004b.
\newblock {S}plitting an operator: {A}n algebraic modularity result and its
  applications to logic programming.
\newblock In {\em {L}ogic {P}rogramming, 20th {I}nternational {C}onference,
  {ICLP} 2004, {P}roceedings}. Lecture Notes in Computer Science, vol. 3132.
  Springer, 195--209.

\bibitem[\protect\citeauthoryear{Verbaeten, Denecker, and Schreye}{Verbaeten
  et~al\mbox{.}}{2000}]{sofiemarc}
{\sc Verbaeten, S.}, {\sc Denecker, M.}, {\sc and} {\sc Schreye, D.~D.} 2000.
\newblock Compositionality of normal open logic programs.
\newblock {\em Journal of Logic Programming\/}~{\em 41}, 151--183.

\end{thebibliography}
\end{document}